\documentclass{article}
\usepackage{float}

\usepackage[main, final]{neurips_2026}

\usepackage[utf8]{inputenc} % allow utf-8 input
\usepackage[T1]{fontenc}    % use 8-bit T1 fonts
\usepackage{hyperref}       % hyperlinks
\usepackage{url}            % simple URL typesetting
\usepackage{booktabs}       % professional-quality tables
\usepackage{amsfonts}       % blackboard math symbols
\usepackage{nicefrac}       % compact symbols for 1/2, etc.
\usepackage{microtype}      % microtypography
\usepackage{xcolor}         % colors
\usepackage{graphicx}
\usepackage{wrapfig}   % in preamble
\usepackage{amsmath}
\usepackage{booktabs}
\usepackage{pifont}
\usepackage[most]{tcolorbox}
\usepackage{subcaption}

\newcommand{\cmark}{\ding{51}}
\newcommand{\xmark}{\ding{55}}

\title{Intent-Aware Caching for Efficient LLM Serving}

\author{%
  Kazi Hasan Ibn Arif\\
  Virginia Tech\\
  Blacksburg, VA, USA\\
  \texttt{hasanarif@vt.edu} \\
  \And
  JinYi Yoon\\
  Inha University\\
  Incheon, South Korea\\
  \texttt{jinyiyoon@inha.ac.kr} \\
  \And
  Dimitrios S. Nikolopoulos\\
  Virginia Tech\\
  Blacksburg, VA, USA\\
  \texttt{dsn@vt.edu} \\
  \AND
  Hans Vandierendonck\\
  Queen's University Belfast\\
  Belfast, UK\\
  \texttt{h.vandierendonck@qub.ac.uk} \\
  \And
  Deepu John\\
  University College Dublin\\
  Dublin, Ireland\\
  \texttt{deepu.john@ucd.ie} \\
  \And
  Bo Ji\\
  Virginia Tech\\
  Blacksburg, VA, USA\\
  \texttt{boji@vt.edu} \\
}

\begin{document}

\maketitle

% \jiny{let's think about the name of our proposed method}
% \hasan{Probably a simple one - IAC - Intent-aware Caching}

% \begin{itemize}
%     \item \hasan{I will replace the figures with pdf for high res later}
% \end{itemize}

% \jiny{a brief explanation of what prefix caching is and then mention this is a key optimization technique}
\begin{abstract}
Modern LLM serving systems use prefix caching to accelerate multi-turn conversational workloads. They cache the key-value states of previous conversation prefixes and reuse them when subsequent requests share the same prefix, to avoid redundant prefill computation and reduce serving latency. However, achieving a high cache hit rate remains challenging because cache reuse depends on whether a conversation continues and how soon the next turn arrives. Our analysis of real workloads shows that user intent is a useful signal for this behavior, as different intents exhibit different continuation probabilities and inter-turn delays. Based on this key insight, we propose Intent-Aware Caching (IAC), an online eviction policy that classifies requests by intent, learns lightweight reuse statistics from recent serving logs and leverages these statistics to guide an oracle-inspired eviction strategy. Extensive evaluation on real workloads (WildChat and ShareChat) shows that IAC reaches a 34.6\% cache hit rate. This substantially outperforms both the previous learning-based approach
%\bo{don't use citation in the abstract}
(18.8\%) and traditional methods such as least recently used (LRU, 12.2\%), while reducing average serving latency by up to 46.1\%. Code is available at \url{https://github.com/hasanar1f/intent-aware-caching}.
% \bo{where is the typical place to mention code?} here 

%
\end{abstract}
%\jiny{maybe something like IntentCache?}
%\hasan{Good one but I feel like it's a bit misleading, sounds like we are caching the intent}
%\jiny{Hmm.. I understand what you mean. Let us think a bit more. Similar to LRU, LPC, etc, IAC seems fine, but if we can name more directly, that will be also good}

% \jiny{Please send me an email when you address all the comments}

% The submitted paper content
% Paper references
% Appendices containing any other supporting textual information e.g. proofs, experimental details (code and data will be submitted in a separate file).
% The NeurIPS paper checklist

\section{Introduction}

%\jiny{Why don't we write the outline and then revise the introduction again? Overall, the main challenges and claims are not well delivered. Also, it is hard to follow if we do not have background knowledge. Finally, the description on our design is too much high-level. Please also add the main contributions.}

\begin{figure*}[t]
  \centering

  \begin{subfigure}[t]{0.4\textwidth}
    \centering
    \includegraphics[width=\linewidth]{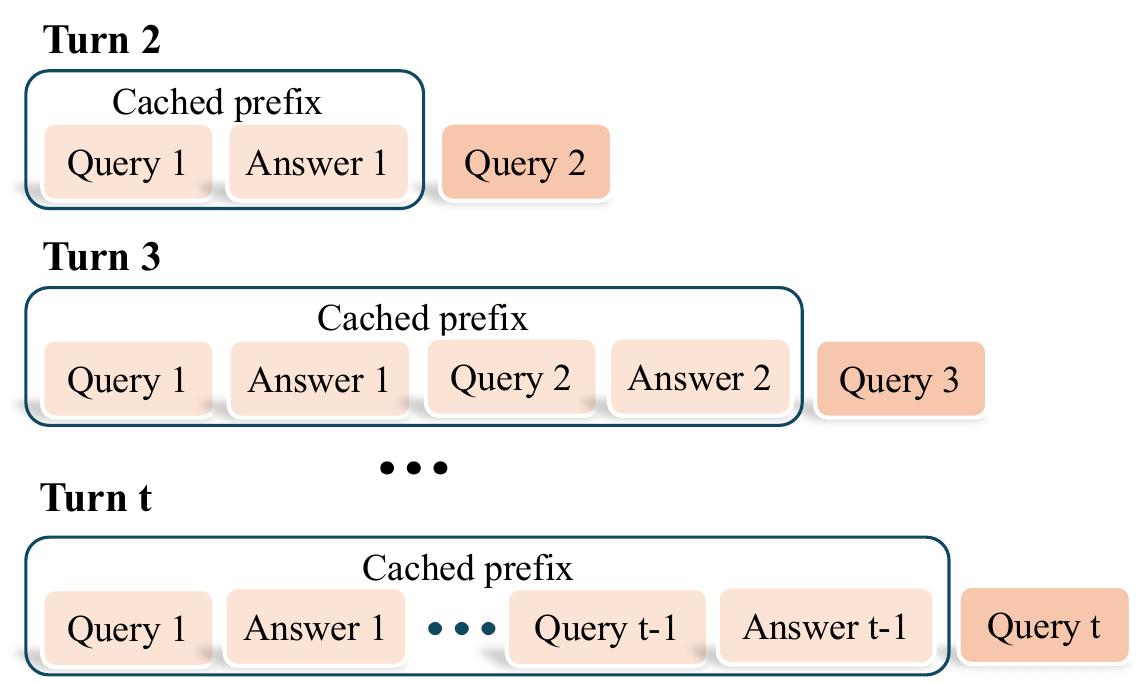}
    \caption{Prefix caching in multi-turn conv.
    %\bo{for the last one, better to use Turn $t$ instead of $t+1$}
    %\bo{you use $t$ to denote the turn index in the main text; be consistent}
    %\bo{to be consistent, use $n$ instead of $N$}
    %\jiny{Is `system prompt' important in our design? If not, just turns would be enough.}
    }
    \label{fig:multi_turn_conv}
  \end{subfigure}
  \hspace{2mm}
  \begin{subfigure}[t]{0.22\textwidth}
    \centering
    \includegraphics[width=\linewidth]{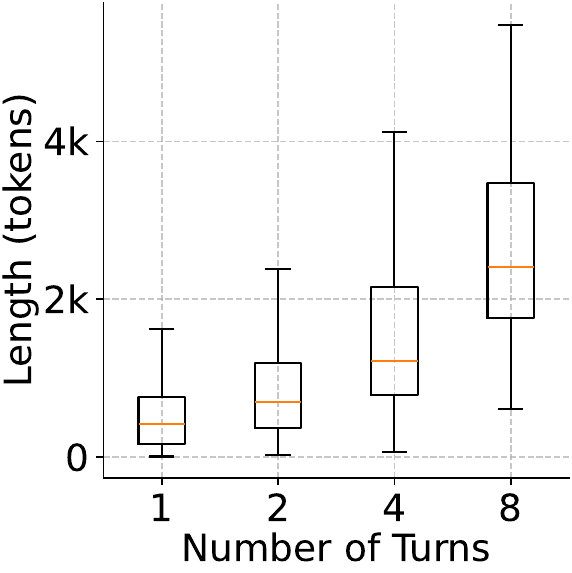}
    \caption{Context len. in turns
    %\jiny{add grid}
    }
    \label{fig:length_distribtion}
  \end{subfigure}
  \hspace{2mm}
  \begin{subfigure}[t]{0.32\textwidth}
    \centering
    \includegraphics[width=\linewidth]{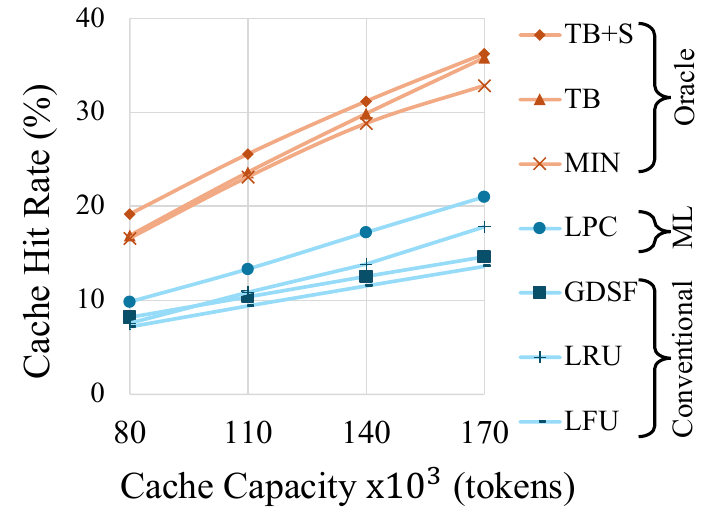}
    \caption{Online policies vs. oracles
    %\bo{use ``Cache Hit Rate (\%)'' for the y axis and remove \% in 40, 30, etc. Same for the other figures involving cache hit rate}
    %\bo{you use \% in the main text; better to keep them consistent}
    %\bo{definite cache hit rate clearly; otherwise, it might be confusing why it is below 0.5}   %\jiny{add x-label}\jiny{by bo--can we show that this is NP-complete even for oracle?}
    }
    \label{fig:gap}
  \end{subfigure}

    \caption{
    (a) In a typical multi-turn conversation, each new request includes previous user queries and model responses as a prefix; caching this prefix reduces redundant prefill computation and lowers TTFT.
    (b) The input context grows with the number of turns, increasing serving cost.
    (c) Both conventional and ML-based policies achieve much lower cache hit rates than oracle policies.
    %\bo{define it in the main text} ... ok moved to main text
    %(i.e., the number of input tokens whose KV states are reused from the cache, divided by the total number of input tokens over all requests).
    %\bo{the meaning of processed is unclear, does it include matched? It's still unclear why the hit rate is upper bounded by 0.5} tokens) than oracle policies, motivating us to close the gap in this work.
     %\jiny{caption? If there is no caption, then separate the subfigures into figures}
  }
  
  \label{fig:workload_characteristics}
  \vspace{-1em}
\end{figure*}

Chatbots using large language models (LLMs), such as ChatGPT~\cite{chatgpt2026} and Gemini~\cite{gemini2026}, serve millions of users every day~\cite{chatterji2025people}. Users often interact with these chatbots through a sequence of follow-up turns to complete different tasks~\cite{wen2025clicks}, commonly known as multi-turn conversations~(Fig.~\ref{fig:multi_turn_conv}). In each turn, the request includes the past conversation history as a prefix, followed by the new user query. As a result, the input context grows with the number of turns~(Fig.~\ref{fig:length_distribtion}), making inference increasingly expensive and resulting in higher time-to-first-token (TTFT) latency and lower throughput (i.e., the total number of tokens processed per second). To serve such prefix-dominated workloads efficiently, modern LLM serving systems adopt \emph{prefix caching}~\cite{MLSYS2024_a66caa17,10.5555/3691992.3691999,openai_prompt_cache}. This is an exact caching mechanism that stores the key-value (KV) states of previously processed tokens and reuses them when subsequent requests share the same prefix, regardless of conversation or user. Therefore, the server only computes the KV states for the newly added tokens instead of recomputing the entire conversation history. Prefix caching directly reduces redundant prefill computation and improves TTFT and throughput~\cite{MLSYS2024_a66caa17,298501}. Accordingly, major providers charge \(10\times\) less for cached input tokens than for uncached ones~\cite{anthropic_pricing,openai_pricing}.

% \hasan{jiny, the re-wrote the whole paragraph}

However, achieving a high cache hit rate~(i.e., the number of input tokens whose KV states are reused from the cache, divided by the total number of input tokens over all requests) remains challenging in existing LLM serving systems, as their eviction policies are typically not designed for conversational workloads~\cite{vllm, sglang}, which differ significantly from traditional object caching. This is because a cached prefix is useful only when the user continues the conversation with a follow-up turn. Even when a follow-up occurs, the reuse delay can vary substantially as users read, think, or compose their next message. In addition, cache reuse is structurally dependent across turns: a hit at turn-\(t\) requires preserving the prefix formed by turns \(1\) through \(t-1\). Cache entries also grow with context length, resulting in highly variable object sizes and making eviction decisions more sophisticated than simple policies used in traditional caching regimes~\cite{cherkasova1998improving,246346}. These challenges are further amplified by workload drift, where cache reuse patterns can shift even within an hour in practical workloads~\cite{wang2025kvcache}.

Conventional policies such as least recently used (LRU), least frequently used (LFU), and greedy dual size-frequency (GDSF) are widely used because they are simple and introduce negligible runtime overhead~\cite{vllm,sglang,wang2025kvcache}. However, they rely only on generic signals such as recency, frequency, and object size. These signals do not directly capture whether a conversation will continue or when its prefix will be reused, resulting in poor eviction decisions. 
%Workload-aware (WA) caching extends GDSF by incorporating learned reuse statistics across request type and turn index~\cite{wang2025kvcache}. However, its four request types (i.e., text, file, search, and image) are too coarse to distinguish diverse conversational reuse patterns, yielding marginal improvement over conventional policies.
Recently proposed Learned Prefix Caching (LPC) trains a machine learning (ML) model to predict the likelihood of follow-up turns from conversation history~\cite{yanglearned}. However, its gains are limited because continuation probability alone does not capture cache reuse patterns~(e.g., a session may continue after a long delay), making the retained prefix less useful under limited GPU memory. Moreover, LLM workloads could drift significantly within an hour~\cite{wang2025kvcache}, making frequently retrained ML predictors operationally expensive.

Offline oracles, in contrast to online policies, have access to future arrival information and therefore provide an upper-bound reference for cache performance. Belady's MIN is optimal when objects have the same size~\cite{belady1966study}. Prefix caches, however, contain KV entries with variable sizes, making the optimal offline eviction problem NP-complete~\cite{chrobak2012caching}. Therefore, prior work on variable-sized object caching proposes practical oracle approximations such as Threshold-Belady (TB) and its size-aware variant TB+S~\cite{246346}. These policies are well suited for prefix caching because cached prefixes differ in both reuse time and size. As shown in Fig.~\ref{fig:gap}, existing online policies remain substantially below oracle performance on real conversational workloads.\footnote{To the best of our knowledge, WildChat~\cite{zhao2024wildchat} and ShareChat~\cite{yan2025sharechat} are the only two public datasets that contain both conversation content and timestamps, comprising a total of 1.2M conversations between real users and ChatGPT-family~models~\cite{chatgpt2026}.} This persistent gap motivates a key question: \emph{How to design an online cache eviction policy for conversational workloads that closes the gap to oracle performance, remains efficient at runtime, and requires low maintenance in deployment?}

%\jiny{add a bit more detail? such as an online eviction policy based on the user intent from the query?}.

In response, we propose \emph{Intent-Aware Caching (IAC)}, an online cache eviction policy that uses workload-specific statistics to better estimate future cache reuse and guide eviction decisions. A key insight behind IAC is that a user's goal in a conversation, which we refer to as the \emph{intent}, reveals useful \emph{cache reuse patterns}. These cache reuse patterns include how many turns a conversation is likely to have (i.e., the number of turns) and how long users tend to wait between turns (i.e., reuse delay). Our experiments on real workloads reveal that grouping requests by intent and turn exposes meaningful patterns in cache reuse (see details in Section~\ref{sec:key_insights}). For example, a conversation about generating code from a user instruction (i.e., \texttt{code\_generation}) often involves more turns than a request for \texttt{translation}.
%\bo{something wroing here; be more careful as this is the camera-ready version}
Similarly, we often observe longer delays for \texttt{code\_generation} tasks than for simple \texttt{chitchat}. Based on this observation,
%\bo{What do you mean by statistical learning here? It is a special term meaning something specific. Seems different from yours.}
IAC estimates cache reuse patterns for each workload category from simple online statistics (i.e., empirical continuation frequencies and a fitted delay distribution) and translates the results into an oracle-inspired eviction policy. IAC consists of three main components (see Fig.~\ref{fig:overview}). First, the lightweight \emph{intent classifier} assigns each request to a workload category. Second, the \emph{statistics estimation} module periodically estimates statistics for each category, including the follow-up likelihood and the delay distribution from recently completed request logs. Finally, the \emph{cache eviction policy} uses these learned statistical signals to mimic
%\bo{use the acronym as you already defined it earlier}size-aware threshold-Belady~\cite{246346}
TB+S, a size-aware oracle policy, to make an eviction decision in an online fashion.

%\jiny{use the same number of floating points}
%\jiny{may add the specific values as well}
%\bo{this para is an elaboration of evaluation results. Shall we move it to after Contributions and give a heading like ``Evaluations.''?}

% \clearpage

\textbf{Contributions.} In this work, we make three main contributions: (i)~we characterize multi-turn conversational workloads through user intent from a cache-reuse perspective, showing that different intents exhibit distinct continuation and inter-turn delay patterns, which are useful for cache management; (ii)~by leveraging these intent-aware cache reuse patterns, we propose IAC, an online cache eviction framework that combines lightweight statistics estimation with an oracle-inspired eviction strategy; (iii)~we implement IAC and evaluate it on realistic setups, demonstrating consistent improvements in cache hit rate, TTFT, and throughput across workloads and model scales.

\textbf{Evaluations.} We implement IAC on SGLang~\cite{sglang}, a widely used LLM serving framework, and evaluate it on real conversational workloads (WildChat~\cite{zhao2024wildchat} and ShareChat~\cite{yan2025sharechat}) using LLMs from 8B to 110B on a realistic multi-GPU setup with 4$\times$A100 (80\,GB) GPUs. Across workloads, model sizes, and cache capacities, IAC consistently outperforms conventional and ML-based policies, narrowing the gap to the oracle. On ShareChat at 140K token capacity, IAC reaches 34.6\% cache hit rate, compared with 12.2\% for LRU and 18.8\% for LPC. These cache gains translate into lower serving latency: on Qwen-1.5-110B, IAC reduces TTFT to 3.76\,s, compared with 6.98\,s for LRU and 5.20\,s for LPC. IAC also sustains higher throughput under load: at 10 requests/s, it reduces average latency to 3.41\,s, compared with 7.01\,s for LRU and 5.44\,s for LPC. Finally, IAC improves memory efficiency: on ShareChat with Llama-3-70B, it reaches a higher cache hit rate with only 24.4\,GB of KV cache than LPC does with 51.9\,GB, saving 53\% of GPU memory. 

%\jiny{So what are the key challenges; why we do experiments in the next paragraph}

% \jiny{summary of main contributions?}

%\jiny{started reading from here}
%\bo{the subsections are quite short; consider other formats}

\vspace{-10pt}
\section{Background and Related Work}
\vspace{-7pt}

%\jiny{Just wondering how it is quadratic} \hasan{attention computation is quadretic and memory is linear}
\textbf{LLM serving for multi-turn conversational workloads.}
LLM inference consists of two phases: \emph{prefill} and \emph{decoding}. Let $x_t$ denote the input request at turn $t$, with a context length of $L_t$ tokens. During prefill, the model processes all $L_t$ input tokens through the transformer layers~\cite{vaswani2017attention} to construct the KV states and produce the first output token.
%\bo{no need to use subscript $l$} layers and hidden dimension $d$, the self-attention computation in prefill has time complexity $O(n_lL_t^2d)$ and KV-state memory cost $O(n_lL_td)$
For a model with $n$ layers and hidden dimension $d$, the self-attention computation in prefill has time complexity $O(nL_t^2d)$ and KV-state memory cost $O(nL_td)$~\cite{vaswani2017attention}. Therefore, the TTFT latency is highly sensitive to the input context length. This becomes particularly costly in multi-turn conversations, where each new request includes the entire conversation history as a prefix (see Fig.~\ref{fig:multi_turn_conv}). Specifically, a turn-$t$ request can be written as $x_t=[u_1,y_1,\ldots,u_{t-1},y_{t-1},u_t]$, where $u_i$ and $y_i$ denote the user query and model response at turn $i$. %\bo{the meaning of the absolute value is not clearly defined here}
Hence, $L_t=\sum_{i=1}^{t-1}(|u_i|+|y_i|)+|u_t|$, where $|\cdot|$ denotes the number of tokens in a sequence, grows with the number of turns. After prefill, decoding reuses the KV states and generates the response autoregressively, requiring one forward pass per generated token. Thus, while both phases contribute to serving cost, multi-turn conversational workloads are especially expensive because the growing conversation history repeatedly increases the prefill cost and directly inflates TTFT latency.

%\jiny{transition: we end the previous paragraph with long prefilling; then the next paragraph should be connected that the prefill is much more expensive and matters in multi-turn conversation}
%\jiny{highly repetitive with explanation in the introduction section}
%\jiny{so what? do we have to do something for prefill in multi-turn settings.. Let's wrap up each section with something related to ours}
%\jiny{Let's highlight the main sentences (the first challenges, the second, etc.). I did some, but please revise if needed. Overall, it should be clearly shown what the main challenges (e.g., using emph or separating the paragraphs) are. Not only here but across the whole paper, it is required to emphasize the main challenges or key claims clearly.}

\textbf{Prefix caching.}
Prefix caching is an exact caching mechanism that stores the KV states of previously processed tokens and reuses them when subsequent requests share the same token prefix, without changing model outputs~\cite{MLSYS2024_a66caa17,openai_prompt_cache,298501}. Formally, for a turn-$t$ request $x_t=[p_t;s_t]$ with prefix $p_t$ of length $|p_t|$ and suffix $s_t$ of length $L_t-|p_t|$, if the KV states of $p_t$ are already cached, the serving system can skip recomputing the prefix and run prefill only on the remaining suffix. Modern serving systems implement this lookup efficiently using prefix-aware cache indices. For example, SGLang's RadixAttention maintains cached KV states in a radix tree, where the runtime performs longest-prefix matching over the incoming token sequence, reuses the KV nodes on the matched path, and computes only the unmatched suffix~\cite{sglang}. This directly reduces the prefill cost from processing $L_t$ tokens to processing only $L_t-|p_t|$ new tokens, thereby lowering TTFT. Prefix caching is independent of user or conversation identity: a cache hit occurs whenever the token prefix matches exactly, regardless of which user or session generated the request. In practice, however, cross-user KV reuse is rare, and most cache hits come from requests submitted by the same user or the same ongoing conversation~\cite{wang2025kvcache}. Multi-turn conversations are strongly prefix caching friendly because the shared prefix is dominated by the accumulated conversation history, e.g., $[u_1,y_1,\ldots,u_{t-1},y_{t-1}]$, which is repeatedly prepended to subsequent turns as illustrated in Fig.~\ref{fig:multi_turn_conv}. Thus, although the decoding cost remains unchanged, prefix caching substantially improves multi-turn serving by eliminating redundant prefill computation over the repeated conversation history, thereby reducing TTFT and improving serving throughput.

\textbf{Related Work.
%\bo{this is the only subsection; you may want to have a subsection called ``Background''; alternatively, just remove the subsection format and use textbf as well}
}
A large body of work improves LLM serving efficiency through batching and scheduling~\cite{ocra, agrawal2024taming, hotprefix}, memory management~\cite{vllm}, and disaggregated inference~\cite{zhong2024distserve, patel2024splitwise}. Conversational workloads amplify this challenge because growing histories make prefill increasingly expensive and TTFT-critical. Prefix caching addresses this bottleneck by storing and reusing KV states when subsequent requests share the same token prefix~\cite{sglang, openai_prompt_cache, dexter2025llm}. In contrast, semantic caching matches requests by meaning rather than exact tokens, which can change model outputs or reduce accuracy~\cite{bang-2023-gptcache, gill2025meancache, liu2025semantic}. Prefix caching preserves exact outputs, but its benefit is limited by GPU memory capacity. Thus, the design of cache eviction policy becomes a central challenge.

Traditional eviction policies, such as LRU, LFU, and GDSF, use generic signals including recency, frequency, and object size~\cite{cherkasova1998improving}. Learning-based policies, such as LRB~\cite{246346}, 3L-Cache~\cite{zhou20253l}, and GL-Cache~\cite{yang2023gl}, learn reuse patterns from past accesses in content delivery networks~\cite{saroiu2002analysis}, storage systems, and object caches. While these works show the value of learned eviction, they target workloads different from multi-turn conversations and are therefore not directly applicable. A production workload study characterizes LLM serving traces and proposes a Workload-Aware (WA) eviction policy that uses request-type-level reuse estimates~\cite{wang2025kvcache}. More recently, LPC trains an ML model to predict conversation continuation likelihood for eviction decisions~\cite{yanglearned}, while T-LRU targets tail TTFT latency to improve SLO attainment~\cite{zhang2025tailoptimized}. Continuum sets a KV-cache time-to-live for multi-turn agents from tool-call durations~\cite{li2025continuum}. Other systems extend effective cache capacity beyond GPU memory through offloading~\cite{liu2024cachegen, 10.1145/3689031.3696098, cheng2025lmcache}. These approaches are complementary to~our work, which focuses on prefix-cache eviction for conversational workloads under a fixed cache capacity.

%\jiny{Shall we add numbers? How much each turn consumes the memory (we had the paper that we used in Amazon proposal)} - we can, but it's not the main point we are making here, in my opinion
%\bo{use Fig. 1 to elaborate on the explanation?}
%\jiny{Why two requests can share the same prefix is missing}
%\bo{better to use subfigures})
%\jiny{transition? how much those caching is burden; then limited GPU memory; thus, high cache hit rate is challenging}
%\bo{unclear what unique challenges are in prefix caching compared to other caching problems. I recall we discussed dependency and heterogeneous sizes}
% \jiny{If this is the first place to mention workload-specific, we should clearly explain what it is.}
%\jiny{Do we know which policy is actually used?} - all four mentioned polices are used
%\jiny{In the box below, we claim that heuristic policies perform poorly, but it is not well delivered in the previous paragraph}
% \bo{these are not heuristic policies; some of them have performance guarantees. In fact, ML-based policies are heuristic policies. Maybe conventional vs. ML-based}

% \jiny{explain briefly what Oracle policy means} 

% \bo{the problem is not even clear yet} 

% \jiny{rethink the subsection or just remove subsections, and change the section to Key Insights for Intent-Aware Caching? Our approach is intent-aware caching, and this section is to introduce existing caching and derive key insights}

\vspace{-7pt}

\section{Key Insights}
\label{sec:key_insights}
\vspace{-7pt}

%\bo{the scope is unclear: do we consider both APIs and human interactions?} - we only consider multi turn chats (human interaction). Previous study shows, while cache reuse is common in api calls, the problem of low cache hit rate is not as pressing as human interaction. Mainly because of the large and variable delay between the reuse.

%\bo{what about reuse across multiple conversations?} - it may exist but way less common in human interaction than APIs. 

% \bo{unclear why chat content and timestamps are important}\bo{the later explanation should come earlier than this} 

The unique characteristics of conversational workloads and the large cache hit rate gap between existing online eviction policies and oracle policies motivate us to characterize this workload more closely. We hypothesize that conversational workloads contain measurable reuse patterns and that an online policy exploiting these patterns can narrow the gap to the oracle.

Existing cache eviction policies struggle with multi-turn conversational workloads. This is because cache reuse is highly heterogeneous, depending not only on whether a conversation continues, but also on how soon the next turn arrives~\cite{wang2025kvcache, servegen}. To characterize this heterogeneity, we analyze two real-world datasets with timestamps: WildChat~\cite{zhao2024wildchat} and ShareChat~\cite{yan2025sharechat}.

\begin{wrapfigure}{r}{0.66\textwidth}
    \centering
    \vspace{-1em}

    \begin{subfigure}[t]{0.495\linewidth}
        \centering
        \includegraphics[width=\linewidth]{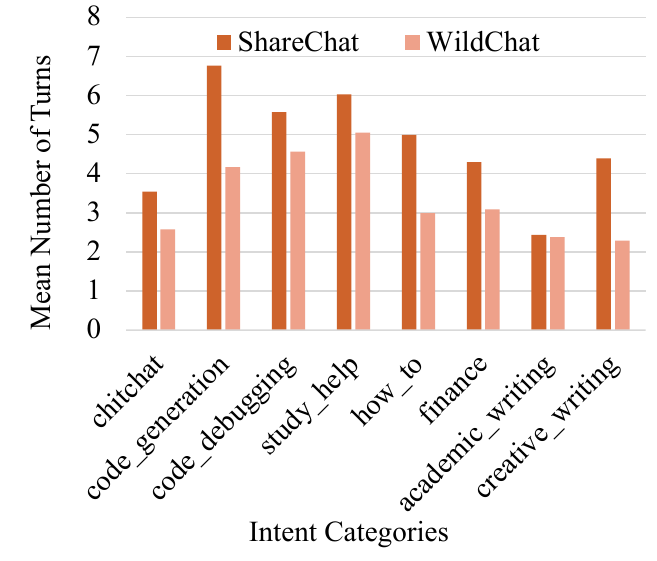}
        \caption{}
        \label{fig:mean_turn_by_intent}
    \end{subfigure}
    \hfill
    \begin{subfigure}[t]{0.495\linewidth}
        \centering
        \includegraphics[width=\linewidth]{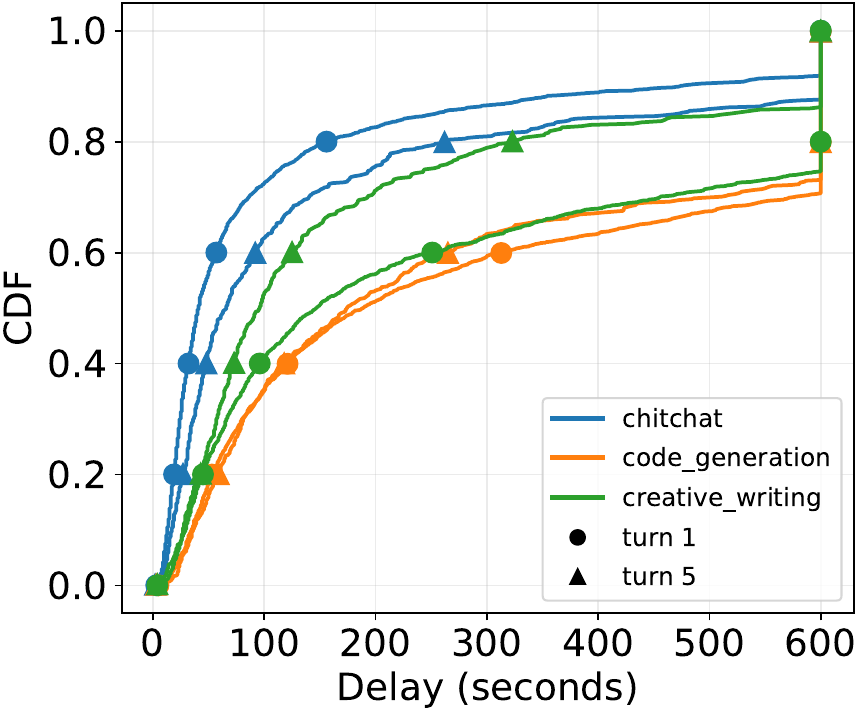}
        \caption{}
        \label{fig:delay_cdf}
    \end{subfigure}

    \caption{
    \textbf{(a)}~\textbf{Insight 1}: Mean number of turns varies across intent categories.
    \textbf{(b)}~\textbf{Insight 2}: Reuse delays vary across intent categories and turns. The delay CDF, clipped at 600\,s, motivates using \((\textit{intent}, \textit{turn})\) as the workload category.
    }
    \vspace{-1em}
    \label{fig:workload_insights}
\end{wrapfigure}

A natural question is which workload signal explains the differences in continuation and delay patterns. We hypothesize that these patterns depend on the user's \emph{intent}, i.e., the task objective expressed by the request~\cite{castillo-lopez-etal-2025-intent}. Different intents impose different interaction structures: for example, code-generation and debugging conversations often require iterative refinement, while travel-planning or information-seeking conversations may terminate after fewer turns. To test this, we assign each request to an intent category using few-shot LLM prompting, following LLM-based topic and intent discovery methods~\cite{10.1145/3637528.3671647} (see details in Section~\ref{sec:intent_classifier} and Appendix~\ref{app:intend_discover_pipeline}). For instance, we label a request as \texttt{code\_generation} when the task is to synthesize code from a specification, as \texttt{code\_debugging} when the task is to diagnose or fix code, and as \texttt{travel\_planning} when the task involves constructing or refining a trip plan. This characterization reveals three key insights as follows.

%\jiny{What are those blue-colored texts in x-axis?} - The numbers looks somewhat cramped and difficult to separate so I colored them. Please suggest any other idea.

% \jiny{Use each ID for each insight to map easily}
% \bo{also, refer to specific figures}
% \jiny{Is this differnet from Insights 1 and 2? It seems a mixture of those two.}\hasan{1-2 are about \# turns while 3 is about delay between turns}
% \bo{I think you need to motivate the importance of intent and turn, followed by an oracle based on such signals, then motivate the online solution}

\textbf{Insight 1: Number of turns varies across intents.}
\label{sec:key_1}
% [REV-OLD] ... whereas requests such as \texttt{how\_to} and \texttt{finance} often end in lower turns.  (Fig. 2a: how_to = 5.0 turns on ShareChat, not low)
In Fig.~\ref{fig:mean_turn_by_intent}, the number of turns in conversations varies across user intents. On ShareChat~\cite{yan2025sharechat}, intents such as \texttt{code\_generation} and \texttt{code\_debugging} tend to involve more turns, whereas requests such as \texttt{academic\_writing} and \texttt{chitchat} often end in fewer turns. We observe the same overall pattern in WildChat~\cite{zhao2024wildchat}. This suggests that some intents naturally create more opportunities for cache reuse than others.

\textbf{Insight 2: Delay varies across intents and turns.}
\label{sec:key_2}
In Fig.~\ref{fig:delay_cdf}, the delay between turns varies across intents and across turns within the same intent category. For example, \texttt{code\_generation} requests exhibit longer delays because users may spend time reading, testing, or modifying the generated code, whereas \texttt{chitchat} tends to have shorter delays. We also observe variations across turn indices within the same intent. For instance, \texttt{creative\_writing} requests at turn 1 and turn 5 show noticeably different delay patterns.
%\bo{do you have an explanation for this difference between different turns?} \hasan{this is just emperical observations, hard to give a direct intuitive explanation}
This suggests that cache reuse depends not only on intent but also on turn. We further analyze the PDFs of delay (see Appendix~\ref{app:additional_insights}), which shows that modeling intent alone already reveals meaningful differences in delay and that combining intent with turn reveals finer-grained reuse patterns.

\begin{wrapfigure}{r}{0.6\textwidth}
    \centering
    \vspace{-1.65em}
    \includegraphics[width=\linewidth]{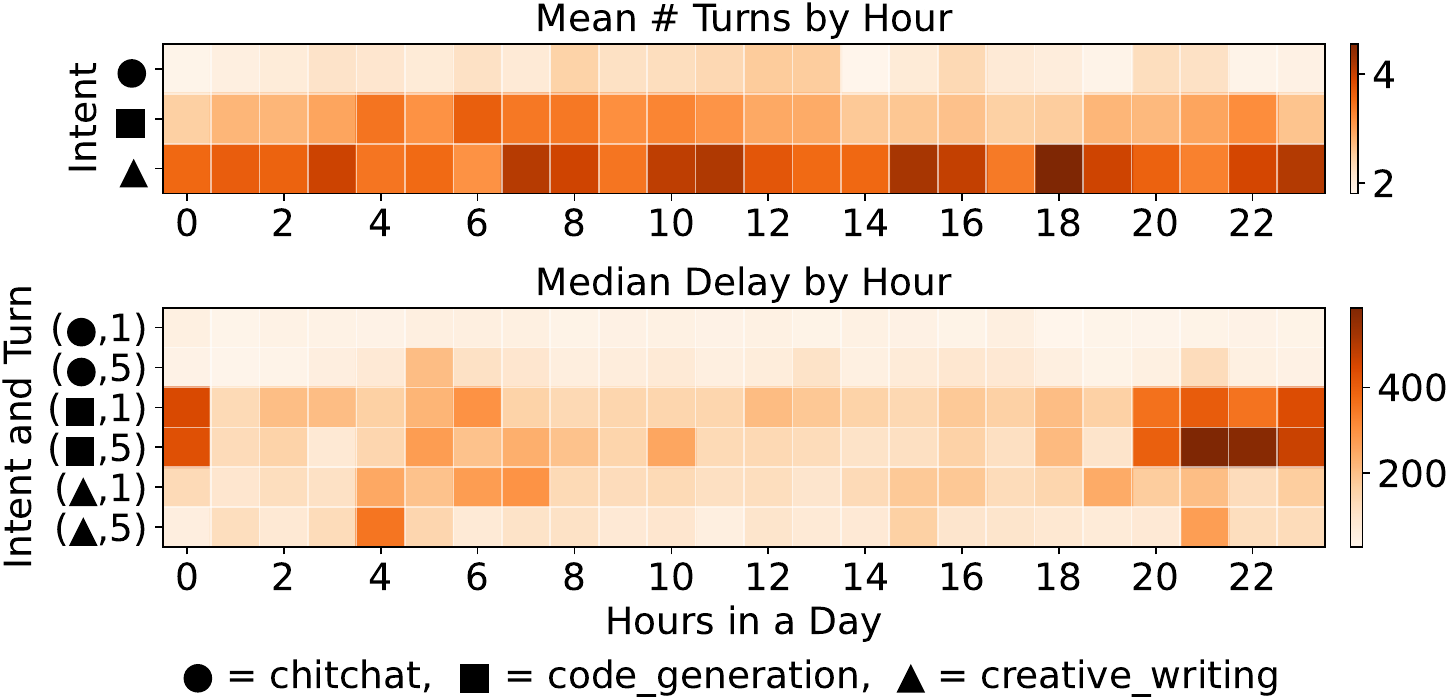}
    
    \caption{\textbf{Insight 3}: Cache reuse patterns (i.e., \#turns and delay) vary over the course of a representative day.}
    \vspace{-1em}
    \label{fig:workload_drift}
\end{wrapfigure}

\textbf{Insight 3: Workload drift.}
\label{sec:key_3}
In Fig.~\ref{fig:workload_drift}, the turn-count and delay patterns across intent categories can shift over the course of a day. For example, on a representative day in WildChat~\cite{zhao2024wildchat}, \texttt{creative\_writing} shows noticeable variation in mean turns, while \texttt{code\_generation} shows variation in delay. This indicates clear workload drift over time. As a result, a static offline-trained ML predictor may not adapt well to such rapid changes without frequent retraining.

% \bo{here, focus on challenges only, not your proposed design}
 
% \bo{is a lightweight intent-classifier hard to design?} 

% \bo{is online statistics estimation hard to do?}

\textbf{Design goal.}
Our goal is to design an online cache eviction policy that is lightweight, adaptive, and effective under the heterogeneous reuse behavior of conversational workloads. This raises three key questions: 
\textbf{(Q1)} How can we identify request intent with minimal overhead? \textbf{(Q2)} How can we continuously learn reuse patterns as workloads drift? \textbf{(Q3)} How can we turn these learned patterns into an effective online cache eviction policy?
% \begin{itemize}
%     \item \textbf{Q1}: How can we identify request intent with minimal overhead? 
%     \item \textbf{Q2}: How can we continuously learn reuse patterns as workloads drift? 
%     \item \textbf{Q3}: How can we turn these learned patterns into an effective online cache eviction policy?
% \end{itemize}

\vspace{-5pt}
\section{Our Design: Intent-Aware Caching (IAC)}
\vspace{-5pt}

% \jiny{provide a high-level overview here}

\begin{wrapfigure}{r}{0.67\textwidth}
    \centering
    \vspace{-1em}
    \includegraphics[width=\linewidth]{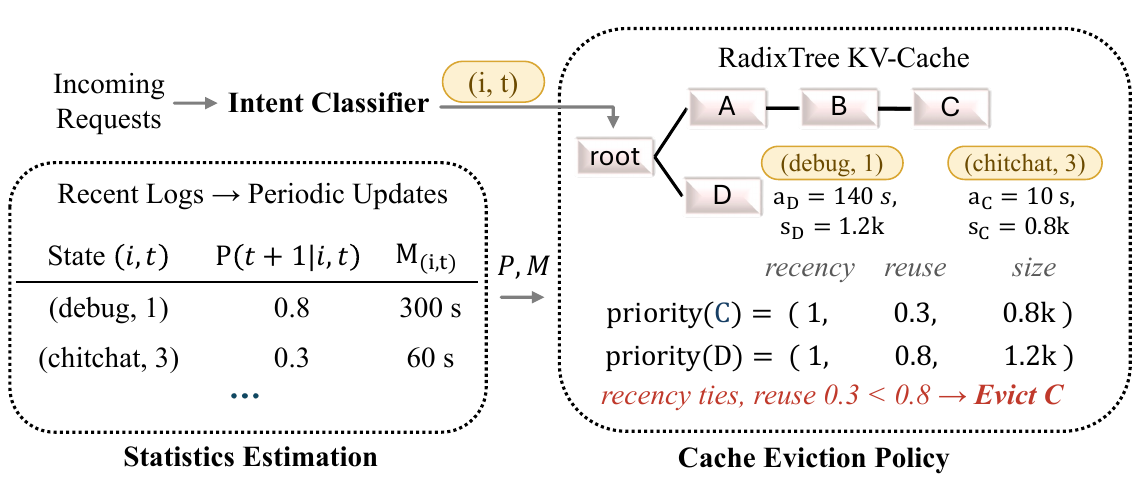}
    \caption{Overview of IAC. The \textbf{Intent Classifier} assigns an intent category to each newly arrived request. The \textbf{Statistics Estimation} module tracks the batch of recent logs and periodically updates cache reuse statistics. During serving, the \textbf{Cache Eviction Policy} (see Eq.~\eqref{eq:iac_score}) computes a priority score for the leaf nodes~(e.g., C and D in the figure) and evicts the node with the lowest score.
    %\bo{what are F and G in the figure?}
    %\bo{the location of this figure needs to be changed - it is an overview, not just for Section 4.2}
    }
    \label{fig:overview}
    \vspace{-1em}
\end{wrapfigure}

By addressing the above questions, we propose \textit{Intent-Aware Caching} (IAC), which consists of three main components~(Fig.~\ref{fig:overview}). First, we identify the request category using a lightweight intent classifier that is suitable for latency-sensitive serving environments (Section~\ref{sec:intent_classifier}). Second, we periodically learn cache reuse behavior through statistics estimation so that the policy can adapt to workload drift (Section~\ref{sec:online_learning}). Lastly, we incorporate the learned statistics into an oracle-inspired online eviction policy (Section~\ref{sec:eviction_policy}).

% \vspace{-5pt}
% \jiny{should be subsection?}
\subsection{Intent Classifier}
\label{sec:intent_classifier}
%\jiny{Most of the contents in this subsection are more about data creation not classifier. Do we have to explain all those datasets here? We can simply mention that we construct the separate training set (see ...). The important thing is how to train.}
% \vspace{-5pt}

\begin{figure}
    \centering
    \includegraphics[width=0.75\linewidth]{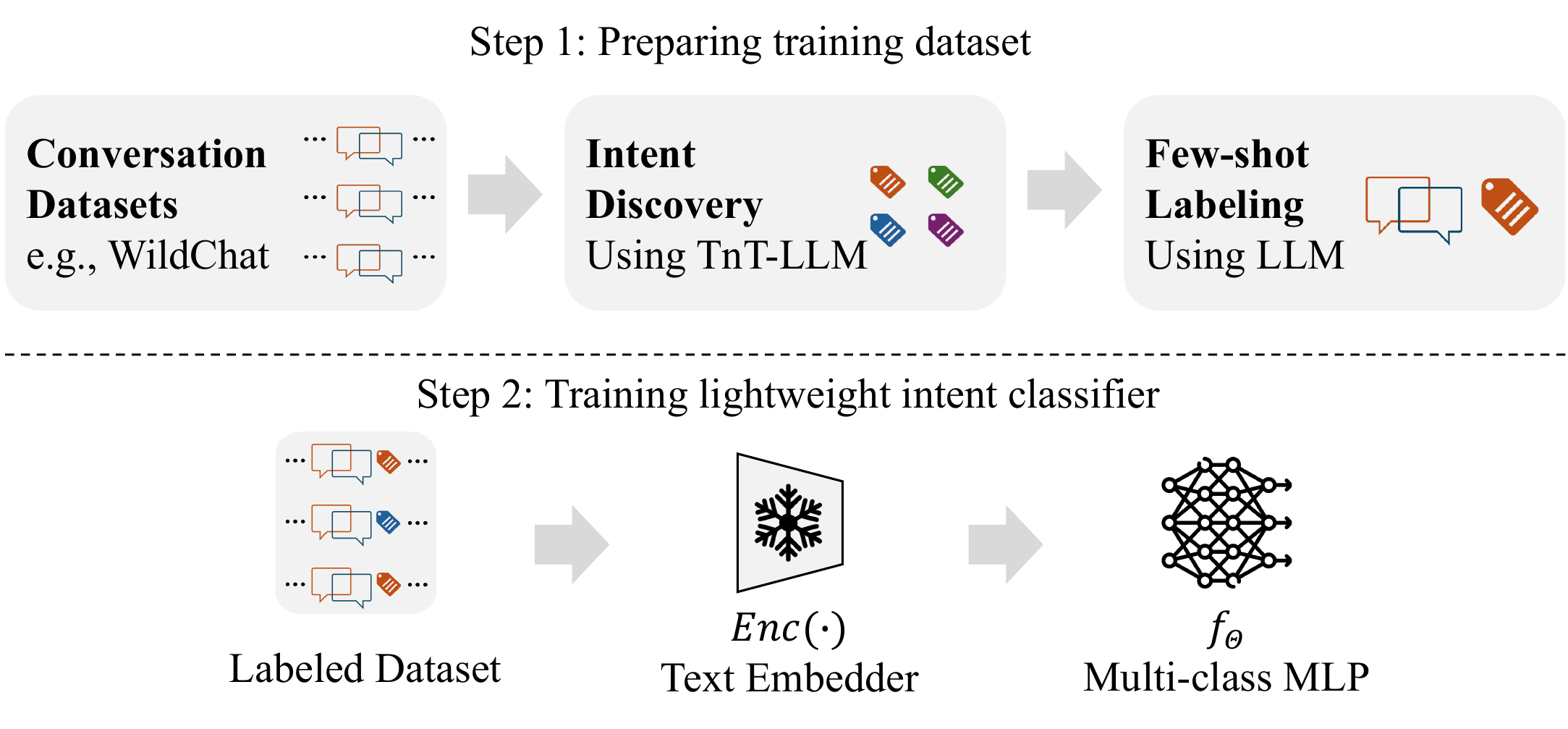}
    \caption{
    Offline training pipeline of the intent classifier. \textbf{Step~1}: intent discovery with TnT-LLM~\cite{10.1145/3637528.3671647} and few-shot LLM labeling. \textbf{Step~2}: training an MLP head on a frozen text embedder.
    %Training pipeline of the intent classifier. In \textbf{Step~1}, we build a labeled training set: we discover an intent taxonomy from conversation datasets with TnT-LLM~\cite{10.1145/3637528.3671647} and label each conversation via few-shot LLM prompting. In \textbf{Step~2}, we train a lightweight multi-class MLP on top of a frozen text embedder. Both steps are performed offline once. During serving, the trained classifier is used to assign an intent category to each incoming request.
    %\bo{which step? You have Step 1 and Step 2 in the figure. Do you mean both steps together?}
    %\bo{try to refine the caption here to better align with the figure}
    %\jiny{It may be misleading that we propose `online' method, but it seems offline. Also, the overall process is not well delivered}} - this step is not part of the design. This is just the training of classifier
    %\jiny{Is this figure important?} \hasan{We can keep this in Appendix}
    }
    \label{fig:training}
    \vspace{-1em}
\end{figure}

To classify a request by workload category at the \((\textit{intent}, \textit{turn})\) level, we need to infer its intent with minimal overhead. Using an LLM or a large ML model defeats this purpose due to additional computational overhead and significantly increased latency during serving. This motivates us to design a separate lightweight intent classifier that can be trained offline once and then reused during deployment. As shown in Fig.~\ref{fig:training}, we build the classifier in two steps: we first construct a labeled dataset by discovering an intent taxonomy and labeling conversations with an LLM (Step~1), and then train a lightweight classifier on this dataset (Step~2).

%\bo{add a sentence to overview the design of intent classifier along with Fig. 5}

%\bo{try to refine the following two paragraphs so that the descriptions are better aligned with Fig.~4.}
%To prepare the training data,

\textbf{Step 1: Preparing the training dataset.} We sample a 200K-conversation split from WildChat~\cite{zhao2024wildchat}, ShareChat~\cite{yan2025sharechat}, ShareGPT~\cite{sharegpt}, and LMSys-Chat~\cite{zheng2023lmsys} to cover diverse domains and languages. This split is strictly disjoint from the online evaluation traces, avoiding data leakage between classifier training and cache-policy evaluation. We then perform \emph{intent discovery} on this split with TnT-LLM, an LLM-based topic discovery framework~\cite{10.1145/3637528.3671647}, using the open-weight GPT-OSS-120B model~\cite{agarwal2025gpt} as the LLM backbone. Starting from the 7 classes introduced in the ShareChat paper~\cite{yan2025sharechat}, it iteratively adds classes to improve coverage. We then remove low-frequency classes and merge semantically similar ones, resulting in 30 intent categories (see Appendix~\ref{app:intend_discover_pipeline} for details). Finally, using this taxonomy, we label the training conversations with few-shot LLM prompting.

\textbf{Step 2: Training a lightweight intent classifier.} After labeling, we train a lightweight intent classifier. The key challenge is to keep the classifier accurate while maintaining millisecond-level inference latency. We therefore pair a \emph{frozen text embedder} \(\mathrm{Enc}\) with a small \emph{multi-class multilayer perceptron (MLP)} head \(f_{\theta}\). Since user requests span diverse domains and languages, we compare several multilingual embedding backbones by accuracy and latency (see Table~\ref{tab:embedding_model_ablation}). Formally, for a turn-\(t\) request, the classifier takes up to the last four user queries \(\tilde{x}_t=[u_{\max(1,t-3)},\ldots,u_t]\), mean-pools their embeddings, and predicts the intent \(\hat{i}=\arg\max_i f_{\theta}(\mathrm{Pool}(\mathrm{Enc}(\tilde{x}_t)))_i\). We keep \(\mathrm{Enc}\) frozen and train only \(f_{\theta}\), a three-layer MLP (embedding dim.\ \(\!\rightarrow\! 2048 \!\rightarrow\! 1024 \!\rightarrow\! 30\)) with a softmax over the 30 intent categories.
%\bo{do you need this notation? use a notation only if you need it, and if so, define it}
%\(d_{\mathrm{emb}}\)-dimensional embedding through hidden layers of 2048 and 1024 units to \(K=30\) intent categories.
We split the labeled data into training, validation, and test sets, train the head with AdamW~\cite{adamw} and a cosine learning-rate schedule, and select the checkpoint with the best validation macro-F1 score (see Appendix~\ref{app:intent_classifier_training} for details).

\vspace{-5pt}

\subsection{Statistics Estimation}
%\hasan{another way}
\label{sec:online_learning}
\vspace{-5pt}

Our analysis in Section~\ref{sec:key_insights} shows that cache reuse depends on both intent and turn. We therefore represent each workload state as a tuple \((i,t)\), where \(i\) denotes the intent category and \(t\) denotes the turn index. To avoid the overhead of frequent retraining, IAC does not learn a predictive ML model online. Instead, it continuously maintains lightweight reuse statistics for each state \((i,t)\). In particular, for every state \((i,t)\), we estimate (i)~the probability that the conversation continues to turn-\((t+1)\)
%\bo{use turn-\((t+1)\) and check other similar places}
and (ii)~the delay distribution until that next turn arrives as follows.

\textbf{Reuse probability.}
For a request in state \((i,t)\), we estimate the probability of receiving a follow-up request at turn-\((t+1)\) as \(P(t+1 \mid i,t) = C(i,t+1) / C(i,t)\), where \(C(i,t)\) denotes the number of intent-\(i\) conversations among the \(N\) most recent conversations that reach turn-\(t\). Since the intent is recomputed at every turn (Section~\ref{sec:intent_classifier}), we assign the most recent intent to the whole conversation once it is completed.
%Thus, \bo{the following sentence is not quite clear}the numerator counts real continuations from state \((i,t)\), not marginal occurrences of intent-\(i\) at turn-\((t+1)\). I dont think we need this line here anyway
The window size \(N\) is a design choice that trades off stability and adaptivity: larger windows provide more stable estimates, while smaller windows adapt faster to workload drift and changing arrival dynamics~(see Appendix~\ref{app:online_serving} for details). 
%{IAC re-estimates the statistics periodically, which takes only about 1.4\,s on CPU off the serving path, and we find \(N=10\)K conversations effective (Section~\ref{sec:ablation}).}

\textbf{Delay distribution.}
For each state \((i,t)\), we model the wall-clock delay between turn-\(t\) and turn-\((t+1)\). Inter-turn delays are typically long-tailed and follow a log-normal distribution~\cite{servegen}. Therefore, we learn \(\log D_{i,t} \sim \mathcal{N}(\mu_{i,t}, \sigma_{i,t}^2)\), where \(\mu_{i,t}\) and \(\sigma_{i,t}^2\) are the mean and variance of the log-delay, respectively. Intuitively, mean \(\mu_{i,t}\) captures the typical delay of state \((i,t)\), while the variance \(\sigma_{i,t}^2\) captures how spread out or variable the delays are. A small \(\mu_{i,t}\) indicates short reuse intervals, and a small \(\sigma_{i,t}\) indicates predictable ones, allowing the eviction policy to distinguish such states from those whose reuse intervals are long or unpredictable.
%\bo{rephrase this to make it better connected to ``short and predictable''}

\vspace{-5pt}

\subsection{Cache Eviction Policy}
\label{sec:eviction_policy}
\vspace{-5pt}

To translate the learned statistics into an online cache eviction rule, we design IAC to mimic the behavior of TB+S%size-aware threshold-Belady (TB+S)
~\cite{246346}, which achieves the best oracle performance on our conversational workloads (Fig.~\ref{fig:gap}). Formally, the oracle priority is \(\left(\mathbf{1}[\Delta \text{next\_arrival} < \text{Threshold}], \text{size}\right)\),
%\bo{is $\mathbf{1}$ the indicator function? If so, need to define it}
where \(\mathbf{1}[\cdot]\) is the indicator function, \(\text{size}\) is the number of tokens, and \(\Delta \text{next\_arrival}\) is the time until the next reuse (set to infinity when no future reuse occurs). %We use \(\text{Threshold}=1000\,\text{s}\), which works well across workloads.
Intuitively, entries that will be reused soon should remain in cache; among entries with the same reuse status, larger entries are preferred for more reusable tokens.

To mimic TB+S's decision online, IAC uses the statistics learned for each workload state \((i,t)\), in the same order as the oracle priority.
%\bo{based on Eq. (1), shouldn't you switch First and Second here?}
First, the recency signal is derived from the inter-turn delay distribution of the state \((i,t)\), which is modeled using its learned parameters. Since the delay is long-tailed, we summarize its central reuse timescale by the median delay obtained from the log-normal model as \(M_{i,t}=\exp(\mu_{i,t})\).
%\bo{how to derive the median from the learned distribution parameters?}
We use this median as the estimated reuse window: if a cached item is younger than \(M_{i,t}\), it is considered to be still within its likely reuse period; otherwise, its chance of near-term reuse is lower. We find the median recency estimate to work better than alternatives such as the mean or mode (Fig.~\ref{fig:mean_meadian_mode_ablation}).
%\bo{\(P(t+1 \mid i,t)\) used here but defined earlier; move the definition here?}
%\bo{shouldn't both age and size depend on $i$ and $t$?} No. age and size are per entry, not per state
%Formally, the priority score \bo{of what?} can be computed as follows:
%\bo{I still think this equation needs to be more precise; it is unclear which age and size they correspond to.}\bo{also, it needs to be consistent with that in Fig. 4, where you have C.priority(...)}
%\bo{this is still not consistent with the figure. In the figure, it's node.priority(). Why don't you use a symbol for a node, let's say $n$. Then $a_n$ is the age of node $n$, $s_n$ is the size, $i_n$ is its intent, $t_n$ is its turn, and priority(n) is its priority. $n$ could be in the subscript or an argument such as $a(n)$. Either should be fine. In the figure, it should be priority(C)<priority(D). Neeed to add the size information for C and D, though}
Second, after this recency test, IAC uses the reuse signal \(P(t+1 \mid i,t)\) estimated in Section~\ref{sec:online_learning}, i.e., the probability that a request in state \((i,t)\) receives a follow-up turn, as the next ordering signal. Finally, it uses size as the last tie-breaker, following TB+S.
%\bo{shouldn't both age and size depend on $i$ and $t$?} No. age and size are per entry, not per state
%Formally, the priority score \bo{of what?} can be computed as follows:
%\bo{I still think this equation needs to be more precise; it is unclear which age and size they correspond to.}\bo{also, it needs to be consistent with that in Fig. 4, where you have C.priority(...)}
%\bo{this is still not consistent with the figure. ... In the figure, it should be priority(C)<priority(D). Neeed to add the size information for C and D, though}
Formally, consider a leaf node \(v\) in the RadixTree, created by a request with intent \(i_v\) at turn \(t_v\). Let \(a_v\) and \(s_v\) be the age (the time since it was cached) and the size (the number of cached tokens) of node \(v\), respectively. Nodes are compared lexicographically (i.e., recency first, followed by reuse, and finally size) by the following priority, and the node with the lowest priority is evicted first:

\vspace{-7pt}

\begin{equation}
\label{eq:iac_score}
\mathrm{priority}(v) =
\bigl(
\underbrace{\mathbf{1}[a_v < M_{i_v,t_v}]}_{\text{Recency signal}},
\;
\underbrace{P(t_v+1 \mid i_v,t_v)}_{\text{Reuse signal}},
\;
s_v
\bigr).
\end{equation}

%\vspace{-7pt}

\section{Evaluation}
\label{sec:evaluation}
%\jiny{we have enough space, so please remove all the vspace commands. Also, if the space is allowed, why don't we include one or two results from the appendix? Maybe Table 8 or Figure 12?} 
\vspace{-7pt}

We evaluate IAC and the baselines along four dimensions: cache hit rate (Section~\ref{sec:cachehitrate}), TTFT latency (Section~\ref{sec:ttft}), throughput under load (Section~\ref{sec:throughput}), and ablation studies (Section~\ref{sec:ablation}).

\textbf{Baselines.}
%\bo{need to mention what the oracle policy is}
We compare IAC with conventional eviction policies (LRU, LFU, and GDSF) and the ML-based LPC policy. In addition, we include TB+S, an offline oracle that knows the next arrival time of every cached entry and therefore serves as an upper-bound reference rather than a deployable policy. In the main comparison, we do not include WA~\cite{wang2025kvcache}, as it relies on reuse statistics over four request types (text, multimodal, file, and search), whereas our primary workloads contain only text requests. Instead, we compare IAC with WA using the cloud trace released by the authors, which includes request-type information (see Appendix~\ref{app:wa_results}). Additionally, we consider a variant of WA adapted for our workloads and evaluate it on these workloads (see Appendix~\ref{app:wa_results}).

\textbf{Implementation and testbed.} We implement IAC in SGLang~\cite{sglang_github} because of its modular design and wide adoption (see Appendix~\ref{app:implementation} for details and Appendix~\ref{app:baselines} for the baselines). Our experiments run on a cloud-hosted multi-GPU node with an Intel Xeon Platinum 8380 CPU, 40 CPU cores, and four NVIDIA A100 (80\,GB) GPUs, connected through PCIe 4.0 $\times$16 links. To evaluate IAC across model scales and deployment settings, we use five open-weight LLMs ranging from 8B to 110B parameters~\cite{grattafiori2024llama, yang2025qwen3, qwen2024qwen15}, which are served on single-, two-, and four-GPU configurations.
%, allowing us to test whether IAC remains effective from small single-GPU serving to larger multi-GPU deployments.

\textbf{Workloads.}
We primarily evaluate IAC on WildChat~\cite{zhao2024wildchat} and ShareChat~\cite{yan2025sharechat}, since both provide original timestamps. Additionally, we use LMSys-Chat~\cite{zheng2023lmsys} and ShareGPT~\cite{sharegpt}, which do not include timestamps, to construct synthetic timestamped workloads and study robustness under different arrival patterns (Section~\ref{sec:cachehitrate}). Finally, we use a production trace from a major cloud provider to compare against WA, since this trace includes the request-type information required by WA~\cite{wang2025kvcache}. For each dataset, after sorting by timestamp, we use the first 40K conversations for LPC training and statistics estimation and use the following 10K conversations to construct the test workload. We provide details of dataset statistics and workload construction in Table~\ref{tab:workloads} and Appendix~\ref{app:experiment_setup}, respectively.

%\textbf{Implementation.}
%We implement IAC on top of SGLang (v0.5.3)~\cite{sglang_github}, an open-source LLM serving system with a modular and extensible runtime. SGLang manages its KV cache using a RadixTree data structure~\cite{sglang}, where each node represents a cached token-prefix segment and shared prefix paths are stored only once across requests. This design enables efficient reuse of KV states for requests with overlapping prefixes (implementation details for IAC and all baselines in Appendix~\ref{app:implementation})

% \vspace{-5pt}

\subsection{Cache Hit Rate}
\label{sec:cachehitrate}
% \vspace{-5pt}

\begin{figure*}[t]
    \centering

    \begin{subfigure}[t]{0.68\textwidth}
        \centering
        \includegraphics[width=\linewidth]{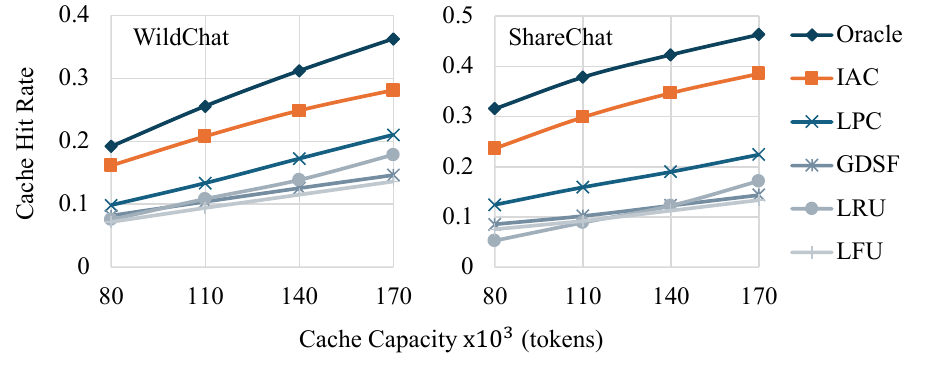}
        \caption{Cache hit rate across various capacities}
        \label{fig:cache_hit_rate}
    \end{subfigure}
    \hfill
    \begin{subfigure}[t]{0.29\textwidth}
        \centering
        \includegraphics[width=\linewidth]{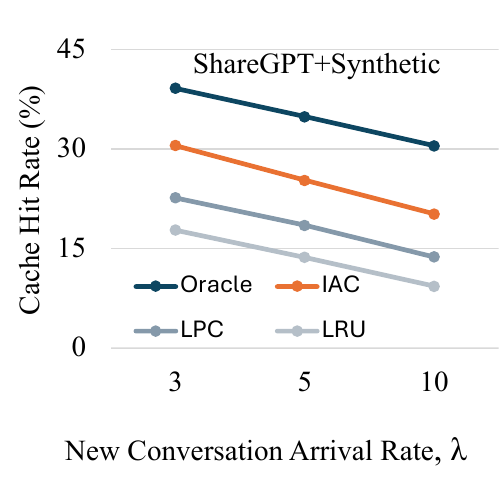}
        \caption{Across arrival rates}
        \label{fig:sharegpt_hit_rate}
    \end{subfigure}

\caption{
    Cache hit rate of IAC vs. the baselines.
    (\textbf{a}) On WildChat and ShareChat, two real workloads with original timestamps, IAC achieves the highest cache hit rate among online policies at every cache capacity.
    (\textbf{b}) On ShareGPT with synthetic Poisson arrival rates \(\lambda\) (new conversations per sec.), IAC consistently outperforms the online baselines even as higher \(\lambda\) increases cache contention.
    }
    \label{fig:cache_hit_eval}
\end{figure*}

Fig.~\ref{fig:cache_hit_rate} shows the cache hit rate on WildChat and ShareChat, two real multi-turn workloads with original timestamps. Across all cache capacities, IAC consistently outperforms both conventional and ML-based baselines. At 140K token capacity (43\,GB for Llama-3-70B) on WildChat, LRU achieves 13.8\%, LPC achieves 17.1\%, and IAC achieves 24.8\% cache hit rate. This corresponds to a 79.7\% improvement over LRU and a 45.0\% improvement over LPC. The gain is larger on ShareChat. At the same capacity, LRU achieves 12.2\%, LPC achieves 18.8\%, and IAC achieves 34.6\%. This corresponds to a 183.6\% improvement over LRU and an 84.0\% improvement over LPC. These results show that IAC substantially improves cache reuse on real conversational workloads.

IAC also improves memory efficiency. On ShareChat with Llama-3-70B, IAC reaches a 23.7\% cache hit rate with only 24.4\,GB of KV cache (80K tokens), whereas LPC reaches only 22.4\% even with 51.9\,GB (170K tokens). This reduces GPU memory demand by 27.5\,GB, corresponding to a 53\% memory saving. 
%\bo{why do you talk about this here, but the figure appears later?} \hasan{I am still taking about fig 6 here. This is just saying high cache hit means memory savings under similar cache hit scenario}
We further evaluate IAC on ShareGPT (Fig.~\ref{fig:sharegpt_hit_rate}) and LMSys-Chat (Appendix~\ref{app:lmsys}) under various arrival patterns while keeping the capacity fixed (150K). For these workloads, new conversations arrive following a Poisson process with rate \(\lambda\), while inter-turn delays are sampled from a fixed long-tailed distribution with mean 4.6 and variance 1, following prior workload studies~\cite{servegen}. The results show that IAC outperforms the baselines and remains robust under various arrival patterns.

\vspace{-5pt}

\subsection{Time-to-First-Token (TTFT) Latency}
\label{sec:ttft}

\vspace{-5pt}

Fig.~\ref{fig:ttft_latency} compares mean TTFT across different LLMs and multi-GPU setups on ShareChat~\cite{yan2025sharechat}, using a request rate of 5 req/s and a fixed cache capacity of 150K tokens. IAC consistently achieves a lower TTFT than both conventional and ML-based policies, with larger gains on bigger models where prefill latency dominates. For example, on Llama-3-70B, IAC reduces TTFT to 1.7\,s, compared with 3.1\,s for LRU, 2.5\,s for LPC, and 1.4\,s for the oracle. On Qwen-1.5-110B, IAC achieves 3.76\,s, compared with 6.98\,s for LRU, 5.20\,s for LPC, and 3.01\,s for the oracle. Overall, IAC reduces TTFT by up to 46.1\% over LRU and up to 32.0\% over LPC, showing that its cache-hit gains translate into substantial latency reductions on real multi-turn workloads. We also report the tail latency in Appendix~\ref{app:tail_latency}.

\vspace{-5pt}

\subsection{Throughput}
\label{sec:throughput}

\vspace{-5pt}

\begin{figure*}[t]
    \centering

    \begin{subfigure}[t]{0.655\textwidth}
        \centering
        \includegraphics[width=\linewidth]{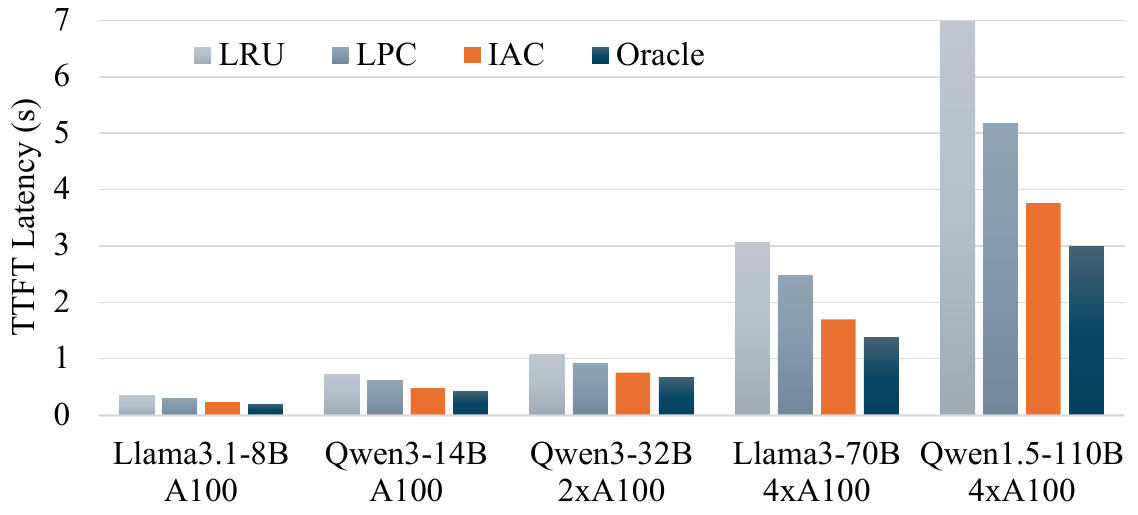}
        \caption{TTFT Latency}
        \label{fig:ttft_latency}
    \end{subfigure}
    \hfill
    \begin{subfigure}[t]{0.335\textwidth}
        \centering
        \includegraphics[width=\linewidth]{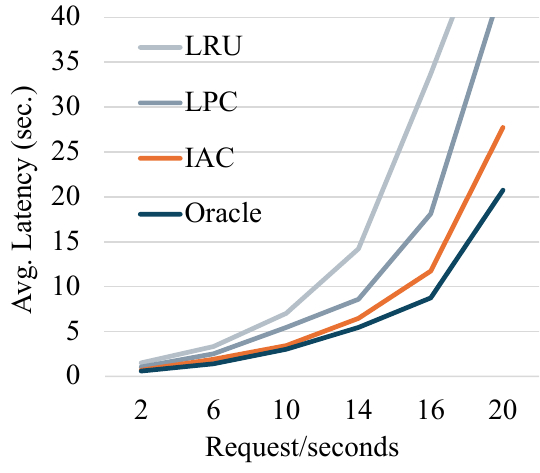}
        \caption{Throughput}
        \label{fig:throughput}
    \end{subfigure}

    \caption{
    (\textbf{a}) IAC reduces TTFT across various models by improving cache hit rate. 
    (\textbf{b}) IAC maintains lower average latency as request rate increases, showing higher throughput under load.
    }
    \label{fig:combined_eval}
\end{figure*}

Fig.~\ref{fig:throughput} reports the average serving latency on Llama-3-70B (4\(\times\)A100). As the load grows, queuing delay increases the average latency of all policies. However, the latency of IAC grows more slowly, because higher cache reuse removes redundant prefill computation and frees GPU time for other requests. For example, at 10 req/s, IAC achieves 3.41\,s latency, compared with 7.01\,s for LRU and 5.44\,s for LPC, reducing latency by 51.4\% and 37.3\%, respectively. Under heavier load at 20 req/s, IAC reduces latency by 49.2\% over LRU and 36.6\% over LPC. Overall, averaged across request rates, IAC reduces serving latency by over 50\% compared with LRU and around 35\% compared with LPC. Equivalently, for a given latency target, IAC can sustain a higher request rate.
%\bo{what do you mean by effective throughput?}  

\vspace{-7pt}

\subsection{Ablation Study}
\label{sec:ablation}

\vspace{-5pt}

\begin{figure*}[t]
    \centering
    \begin{subfigure}[t]{0.39\textwidth}
        \centering
        \includegraphics[width=\linewidth]{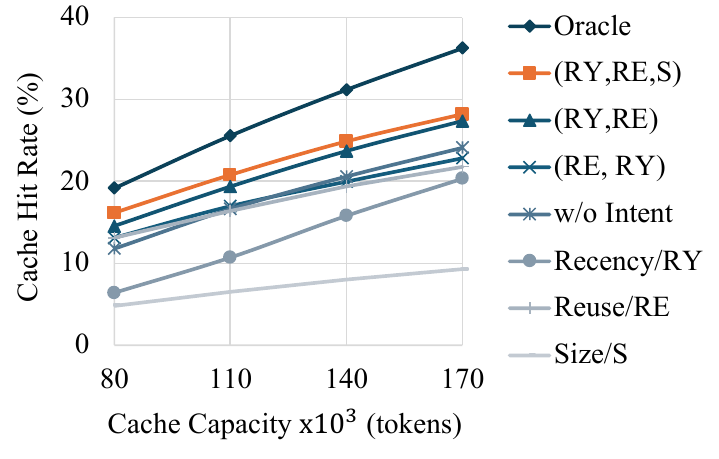}
        \caption{Eviction-policy ablation}
        \label{fig:ablation_signals}
    \end{subfigure}
    \hfill
    \begin{subfigure}[t]{0.33\textwidth}
        \centering
        \includegraphics[width=\linewidth]{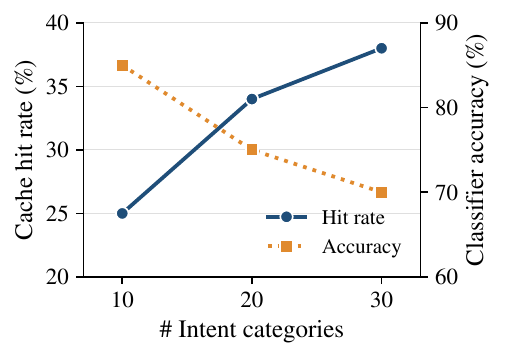}
        \caption{Intent granularity
        %\bo{you may have dual y-axis and two curves, where the right y-axis is for accuracy}
        }
        %\bo{why not make them subfigures (b) and (c)?}
        %\bo{this figure is a bit confusing and hard to read}
        \label{fig:granularity}
    \end{subfigure}
    \hfill
    \begin{subfigure}[t]{0.26\textwidth}
        \centering
        \includegraphics[width=\linewidth]{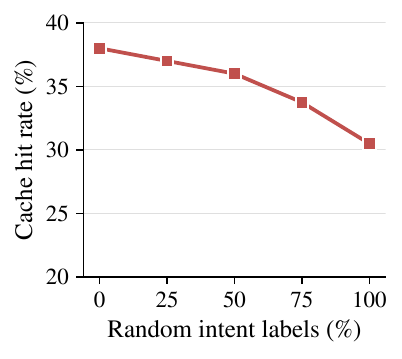}
        \caption{Classifier label noise}
        \label{fig:noise}
    \end{subfigure}

\caption{
        Ablation and sensitivity of IAC.
        (\textbf{a}) Intent-turn grouping with the recency, reuse, and size signals performs best.
        (\textbf{b}) Finer intent categories lower classifier accuracy but raise cache hit rate by preserving reuse differences that coarser groups merge.
        (\textbf{c}) Cache hit rate degrades gracefully as intent labels are replaced with random ones, since the turn, recency, and size signals remain intact.
        %\bo{need to move this figure to  the previous page, closer to where it is first discussed}
    }
    \label{fig:ablation_all}
    \vspace{-1em}
\end{figure*}

%\bo{need to rearrange the figures or the questions; their orders do not match}
\textit{How much does each IAC signal contribute, and does the ordering matter?} As shown in Fig.~\ref{fig:ablation_signals}, we ablate the three signals used by IAC: recency (RY), reuse probability (RE), and size (S). Each signal improves cache hit rate, but reuse is stronger than recency when used alone, showing that follow-up likelihood is the stronger individual predictor of prefix reuse. Combining RY and RE improves performance further, and the ordering matters: $(\mathrm{RY},\mathrm{RE})$ consistently outperforms $(\mathrm{RE},\mathrm{RY})$.
%Since the priority is lexicographic, a continuous RE in \((0,1)\) placed first rarely ties, so the binary RY in \(\{0,1\}\) is rarely used. 
%\bo{if that's the case, how come size is still useful?}
Adding size as the final tie-breaker gives $(\mathrm{RY},\mathrm{RE},\mathrm{S})$, which achieves the best performance among all online variants and remains closest to the oracle across cache capacities.

\textit{Does intent provide additional value beyond turn?} 
To isolate the role of intent, we remove intent from the workload category and group requests only by turn index (referred to as \textit{w/o Intent}) in Fig.~\ref{fig:ablation_signals}. This reduces the cache hit rate, showing that the turn index alone is not sufficient. Combining intent and turn creates more effective groups for learning conversational reuse.

\textit{How many intent categories are needed?} We study the trade-off between performance and intent granularity in Fig.~\ref{fig:granularity}. Our TnT-LLM~\cite{10.1145/3637528.3671647}-based intent discovery produces a 30-category taxonomy. To test whether fewer categories are sufficient, we merge it into coarser 20-category and 10-category variants. We provide the full merge lists in Appendices~\ref{app:intent_20} and \ref{app:intent_10}. Coarser grouping improves classifier accuracy because the classification task becomes easier. However, it also removes behavioral distinctions that matter for caching. As a result, the 10-category variant achieves higher classifier accuracy but much lower cache hit rate, while the 30-category taxonomy gives the best cache performance. This shows that preserving fine-grained reuse structure is more important than maximizing classifier accuracy alone.

\textit{How accurate does the intent classifier need to be?} To test sensitivity to classifier errors, we randomly replace a fraction of predicted intent labels at serving time and measure the cache hit rate at a fixed cache capacity (170K tokens) on ShareChat, presented in Fig.~\ref{fig:noise}. We vary the fraction of randomized intent labels from 0\% to 100\%. Results show that IAC degrades gracefully as classifier errors increase.
%\bo{even with 100\% noise, the cache hit rate is still 30\%? Any explanation?}
Even with fully random labels, IAC still achieves a 30\% cache hit rate because randomizing the intent removes only the intent signal while the other signals remain intact, and each random-intent group becomes a random sample of all conversations at turn \(t\), so its statistics approach turn-level averages. This indicates that IAC does not require a perfect classifier. Instead, it only needs intent predictions to be accurate enough to preserve coarse reuse differences across workload groups.

\begin{wrapfigure}{r}{0.4\textwidth}
    \centering
    \vspace{-1em}
    \includegraphics[width=\linewidth]{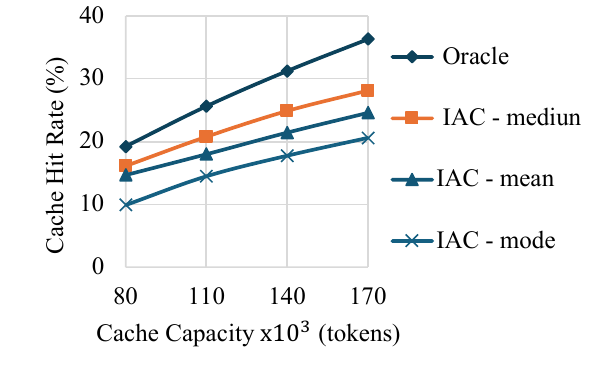}
    \caption{Reuse-window estimators.}
    \vspace{-1em}
    \label{fig:mean_meadian_mode_ablation}
\end{wrapfigure}

\textit{Why do we use the median delay as the reuse window?}
We compare three estimators of the reuse window \(M_{i,t}\) derived from the learned log-normal delay model: the mean \(e^{\mu_{i,t}+\sigma_{i,t}^2/2}\), the median \(e^{\mu_{i,t}}\), and the mode \(e^{\mu_{i,t}-\sigma_{i,t}^2}\). As shown in Fig.~\ref{fig:mean_meadian_mode_ablation}, the median provides the best performance. This is because inter-turn delays in conversational workloads are typically long-tailed (Fig.~\ref{fig:delay_cdf}), where a small fraction of very large delays skews the distribution. The mean is sensitive to these outliers and overestimates the reuse window, causing the policy to retain stale entries longer than useful. The mode, on the other hand, captures only the most frequent short-delay region and underestimates the reuse window, resulting in premature eviction of reusable prefixes. In contrast, the median provides a robust estimate of the typical reuse timescale, balancing early and late reuse, and therefore yields a higher cache hit rate.

\begin{wraptable}{r}{0.38\textwidth}
\centering
\vspace{-1.2em}
\caption{Embedding backbone models.}
\label{tab:embedding_model_ablation}
\small
\setlength{\tabcolsep}{2pt}
\begin{tabular}{@{}lcc@{}}
\toprule
Model (size) & Acc. & Lat. \\
\midrule
ModernBERT (0.2B+3.7M)     & 62\% & 33 ms \\
E5-Small (0.1B+2.9M)       & 70\% & 13 ms \\
E5-Base (0.2B+3.7M)        & 71\% & 16 ms \\
E5-Large (0.5B+4.2M)       & 72\% & 20 ms \\
E5-Large-Inst (0.5B+4.2M) & 75\% & 20 ms \\
\bottomrule
\end{tabular}
\vspace{-1em}
\end{wraptable}

\textit{Which embedding model is sufficient for intent classification?} We compare embedding backbones by classifier accuracy and inference latency in Table~\ref{tab:embedding_model_ablation}. The E5 family~\cite{wang2024multilingual} consistently outperforms ModernBERT~\cite{modernbert} while maintaining low latency, making it a better fit for online serving. This is likely because E5 models are trained for multilingual text embedding, which better matches real chatbot workloads with diverse user queries. Among the E5 variants, the instruction-tuned E5-Large-Inst achieves the best accuracy, reaching 75\% with 20\,ms latency. Therefore, we choose E5-Large-Inst as the backbone in IAC.

\vspace{-10pt}

% \section{Conclusion, Limitations, and Future Work}
\section{Conclusion}
\label{sec:conclusion}

\vspace{-10pt}

This work studies prefix-cache eviction for multi-turn conversational workloads. Our key observation is that cache reuse is highly heterogeneous: different requests vary in both reuse likelihood and inter-turn delay. IAC exploits this heterogeneity by grouping requests based on intent and turn index, and then learning lightweight reuse statistics within each group. The outcome of this study suggests that practical prefix-cache eviction should move beyond generic signals and exploit workload-aware signals that capture how users interact with LLM systems.

While this paper advances prefix-cache eviction for multi-turn conversational workloads, there are still several limitations. First, we evaluate IAC mainly on conversational workloads, where cache reuse is driven by repeated multi-turn context. However, prefix caching can also benefit other workloads with repeated prefixes, such as agentic workflows, tool-use pipelines, and API applications with shared system prompts or templates. Second, our intent classifier keeps the embedding backbone frozen and trains only a lightweight MLP prediction head. The ablation in Table~\ref{tab:embedding_model_ablation} suggests that stronger or task-adapted embeddings may improve intent classification, which could further improve cache eviction decisions and cache hit rate. Future work can consider extensions that include broader workload signals and broader LLM serving settings. Furthermore, workload-aware signals, such as intent, turn structure, and tool usage, could be used beyond cache eviction to jointly optimize batching, scheduling, and routing decisions across model fleets.

\vspace{-5pt}

\section*{Acknowledgments}

\vspace{-10pt}

%Hans:
This work was supported in part by the National Science Foundation under Grant CNS-2315851, the Commonwealth Cyber Initiative, the Virginia Tech College of Engineering, the Virginia Tech Institute for Advanced Computing, a research grant from the Department for the Economy, Northern Ireland, under the US-Ireland R\&D Partnership Programme grant agreement USI-226, Taighde Éireann – Research Ireland (22/US/3848) under the US-Ireland R\&D Partnership Programme, 
%DSN: two internal grants supported, a CoE MGI and an IAC grant
%Jiny:
and the National Research Foundation of Korea (NRF) grant funded by the Korea government (MSIT) (No. RS-2026-25602748).

\bibliographystyle{unsrt}
\bibliography{neurips_2026}

\begin{thebibliography}{10}

\bibitem{chatgpt2026}
{OpenAI}.
\newblock {ChatGPT}.
\newblock \url{https://chatgpt.com}, 2026.
\newblock Conversational AI based on large language models.

\bibitem{gemini2026}
{Google}.
\newblock {Gemini}.
\newblock \url{https://gemini.google.com}, 2026.
\newblock Conversational AI based on large language models.

\bibitem{chatterji2025people}
Aaron Chatterji, Thomas Cunningham, David~J Deming, Zoe Hitzig, Christopher Ong, Carl~Yan Shan, and Kevin Wadman.
\newblock How people use chatgpt.
\newblock Technical report, National Bureau of Economic Research, 2025.

\bibitem{wen2025clicks}
Qifu Wen, Prishita Kochhar, Sherif Zeyada, Tahereh Javaheri, and Reza Rawassizadeh.
\newblock From clicks to conversations: Evaluating the effectiveness of conversational agents in statistical analysis.
\newblock {\em International Journal of Human--Computer Interaction}, pages 1--18, 2025.

\bibitem{MLSYS2024_a66caa17}
In~Gim, Guojun Chen, Seung-seob Lee, Nikhil Sarda, Anurag Khandelwal, and Lin Zhong.
\newblock Prompt cache: Modular attention reuse for low-latency inference.
\newblock In P.~Gibbons, G.~Pekhimenko, and C.~De Sa, editors, {\em Proceedings of Machine Learning and Systems}, volume~6, pages 325--338, 2024.

\bibitem{10.5555/3691992.3691999}
Bin Gao, Zhuomin He, Puru Sharma, Qingxuan Kang, Djordje Jevdjic, Junbo Deng, Xingkun Yang, Zhou Yu, and Pengfei Zuo.
\newblock Cost-efficient large language model serving for multi-turn conversations with cachedattention.
\newblock In {\em Proceedings of the 2024 USENIX Conference on Usenix Annual Technical Conference}, USENIX ATC'24, USA, 2024. USENIX Association.

\bibitem{openai_prompt_cache}
{OpenAI}.
\newblock https://developers.openai.com/api/docs/guides/prompt-caching, 2026.
\newblock Accessed on Mar 8, 2026.

\bibitem{298501}
Bin Gao, Zhuomin He, Puru Sharma, Qingxuan Kang, Djordje Jevdjic, Junbo Deng, Xingkun Yang, Zhou Yu, and Pengfei Zuo.
\newblock {Cost-Efficient} large language model serving for multi-turn conversations with {CachedAttention}.
\newblock In {\em 2024 USENIX Annual Technical Conference (USENIX ATC 24)}, pages 111--126, Santa Clara, CA, July 2024. USENIX Association.

\bibitem{anthropic_pricing}
{Anthropic}.
\newblock https://platform.claude.com/docs/en/about-claude/pricing, 2026.
\newblock Accessed on Mar 8, 2026.

\bibitem{openai_pricing}
{OpenAI}.
\newblock https://openai.com/api/pricing/, 2026.
\newblock Accessed on Mar 8, 2026.

\bibitem{vllm}
Woosuk Kwon, Zhuohan Li, Siyuan Zhuang, Ying Sheng, Lianmin Zheng, Cody~Hao Yu, Joseph Gonzalez, Hao Zhang, and Ion Stoica.
\newblock Efficient memory management for large language model serving with pagedattention.
\newblock In {\em SOSP '23}, page 611–626, New York, NY, USA, 2023. Association for Computing Machinery.

\bibitem{sglang}
Lianmin Zheng, Liangsheng Yin, Zhiqiang Xie, Chuyue Sun, Jeff Huang, Cody~H Yu, Shiyi Cao, Christos Kozyrakis, Ion Stoica, Joseph~E Gonzalez, et~al.
\newblock Sglang: Efficient execution of structured language model programs.
\newblock {\em Advances in neural information processing systems}, 37:62557--62583, 2024.

\bibitem{cherkasova1998improving}
Ludmila Cherkasova.
\newblock {\em Improving WWW proxies performance with greedy-dual-size-frequency caching policy}.
\newblock Hewlett-Packard Laboratories Palo Alto, CA, USA, 1998.

\bibitem{246346}
Zhenyu Song, Daniel~S. Berger, Kai Li, and Wyatt Lloyd.
\newblock Learning relaxed belady for content distribution network caching.
\newblock In {\em 17th USENIX Symposium on Networked Systems Design and Implementation (NSDI 20)}, pages 529--544, Santa Clara, CA, February 2020. USENIX Association.

\bibitem{wang2025kvcache}
Jiahao Wang, Jinbo Han, Xingda Wei, Sijie Shen, Dingyan Zhang, Chenguang Fang, Rong Chen, Wenyuan Yu, and Haibo Chen.
\newblock $\{$KVCache$\}$ cache in the wild: Characterizing and optimizing $\{$KVCache$\}$ cache at a large cloud provider.
\newblock In {\em 2025 USENIX Annual Technical Conference (USENIX ATC 25)}, pages 465--482, 2025.

\bibitem{yanglearned}
Dongsheng Yang, Austin Li, Kai Li, and Wyatt Lloyd.
\newblock Learned prefix caching for efficient llm inference.
\newblock In {\em The Thirty-ninth Annual Conference on Neural Information Processing Systems}, 2025.

\bibitem{belady1966study}
Laszlo~A. Belady.
\newblock A study of replacement algorithms for a virtual-storage computer.
\newblock {\em IBM Systems journal}, 5(2):78--101, 1966.

\bibitem{chrobak2012caching}
Marek Chrobak, Gerhard~J Woeginger, Kazuhisa Makino, and Haifeng Xu.
\newblock Caching is hard—even in the fault model.
\newblock {\em Algorithmica}, 63(4):781--794, 2012.

\bibitem{zhao2024wildchat}
Wenting Zhao, Xiang Ren, Jack Hessel, Claire Cardie, Yejin Choi, and Yuntian Deng.
\newblock Wildchat: 1m chat{GPT} interaction logs in the wild.
\newblock In {\em The Twelfth International Conference on Learning Representations}, 2024.

\bibitem{yan2025sharechat}
Yueru Yan, Tuc Nguyen, Bo~Su, Melissa Lieffers, and Thai Le.
\newblock Sharechat: A dataset of chatbot conversations in the wild.
\newblock {\em arXiv preprint arXiv:2512.17843}, 2025.

\bibitem{vaswani2017attention}
Ashish Vaswani, Noam Shazeer, Niki Parmar, Jakob Uszkoreit, Llion Jones, Aidan~N Gomez, {\L}ukasz Kaiser, and Illia Polosukhin.
\newblock Attention is all you need.
\newblock {\em Advances in neural information processing systems}, 30, 2017.

\bibitem{ocra}
Gyeong-In Yu, Joo~Seong Jeong, Geon-Woo Kim, Soojeong Kim, and Byung-Gon Chun.
\newblock Orca: A distributed serving system for {Transformer-Based} generative models.
\newblock In {\em 16th USENIX Symposium on Operating Systems Design and Implementation (OSDI 22)}, pages 521--538, Carlsbad, CA, July 2022. USENIX Association.

\bibitem{agrawal2024taming}
Amey Agrawal, Nitin Kedia, Ashish Panwar, Jayashree Mohan, Nipun Kwatra, Bhargav Gulavani, Alexey Tumanov, and Ramachandran Ramjee.
\newblock Taming $\{$Throughput-Latency$\}$ tradeoff in $\{$LLM$\}$ inference with $\{$Sarathi-Serve$\}$.
\newblock In {\em 18th USENIX symposium on operating systems design and implementation (OSDI 24)}, pages 117--134, 2024.

\bibitem{hotprefix}
Yuhang Li, Rong Gu, Chengying Huan, Zhibin Wang, Renjie Yao, Chen Tian, and Guihai Chen.
\newblock Hotprefix: Hotness-aware kv cache scheduling for efficient prefix sharing in llm inference systems.
\newblock {\em Proc. ACM Manag. Data}, 3(4), September 2025.

\bibitem{zhong2024distserve}
Yinmin Zhong, Shengyu Liu, Junda Chen, Jianbo Hu, Yibo Zhu, Xuanzhe Liu, Xin Jin, and Hao Zhang.
\newblock $\{$DistServe$\}$: Disaggregating prefill and decoding for goodput-optimized large language model serving.
\newblock In {\em 18th USENIX Symposium on Operating Systems Design and Implementation (OSDI 24)}, pages 193--210, 2024.

\bibitem{patel2024splitwise}
Pratyush Patel, Esha Choukse, Chaojie Zhang, Aashaka Shah, {\'I}{\~n}igo Goiri, Saeed Maleki, and Ricardo Bianchini.
\newblock Splitwise: Efficient generative llm inference using phase splitting.
\newblock In {\em 2024 ACM/IEEE 51st Annual International Symposium on Computer Architecture (ISCA)}, pages 118--132. IEEE, 2024.

\bibitem{dexter2025llm}
Gregory Dexter, Shao Tang, Ata Fatahibaarzi, Qingquan Song, Tejas Dharamsi, and Aman Gupta.
\newblock {LLM} query scheduling with prefix reuse and latency constraints.
\newblock In {\em The Thirty-ninth Annual Conference on Neural Information Processing Systems}, 2025.

\bibitem{bang-2023-gptcache}
Fu~Bang.
\newblock {GPTC}ache: An open-source semantic cache for {LLM} applications enabling faster answers and cost savings.
\newblock In Liling Tan, Dmitrijs Milajevs, Geeticka Chauhan, Jeremy Gwinnup, and Elijah Rippeth, editors, {\em Proceedings of the 3rd Workshop for Natural Language Processing Open Source Software (NLP-OSS 2023)}, pages 212--218, Singapore, December 2023. Association for Computational Linguistics.

\bibitem{gill2025meancache}
Waris Gill, Mohamed Elidrisi, Pallavi Kalapatapu, Ammar Ahmed, Ali Anwar, and Muhammad~Ali Gulzar.
\newblock Meancache: User-centric semantic caching for llm web services.
\newblock In {\em 2025 IEEE International Parallel and Distributed Processing Symposium (IPDPS)}, pages 1298--1310. IEEE, 2025.

\bibitem{liu2025semantic}
Xutong Liu, Baran Atalar, Xiangxiang Dai, Jinhang Zuo, Siwei Wang, John~C.S. Lui, Wei Chen, and Carlee Joe-Wong.
\newblock Semantic caching for low-cost llm serving: From offline learning to online adaptation.
\newblock In {\em IEEE International Conference on Computer Communications (INFOCOM)}, May 2026.

\bibitem{zhou20253l}
Wenbin Zhou, Zhixiong Niu, Yongqiang Xiong, Juan Fang, and Qian Wang.
\newblock $\{$3L-Cache$\}$: Low overhead and precise learning-based eviction policy for caches.
\newblock In {\em 23rd USENIX Conference on File and Storage Technologies (FAST 25)}, pages 237--254, 2025.

\bibitem{yang2023gl}
Juncheng Yang, Ziming Mao, Yao Yue, and KV~Rashmi.
\newblock $\{$GL-Cache$\}$: Group-level learning for efficient and high-performance caching.
\newblock In {\em 21st USENIX Conference on File and Storage Technologies (FAST 23)}, pages 115--134, 2023.

\bibitem{saroiu2002analysis}
Stefan Saroiu, Krishna~P Gummadi, Richard~J Dunn, Steven~D Gribble, and Henry~M Levy.
\newblock An analysis of internet content delivery systems.
\newblock {\em ACM SIGOPS Operating Systems Review}, 36(SI):315--327, 2002.

\bibitem{zhang2025tailoptimized}
Wenxin Zhang, Yueying Li, Ciamac~C. Moallemi, and Tianyi Peng.
\newblock Tail-optimized caching for {LLM} inference.
\newblock In {\em NeurIPS 2025 Workshop MLxOR: Mathematical Foundations and Operational Integration of Machine Learning for Uncertainty-Aware Decision-Making}, 2025.

\bibitem{li2025continuum}
Hanchen Li, Runyuan He, Qiuyang Mang, Qizheng Zhang, Huanzhi Mao, Xiaokun Chen, Hangrui Zhou, Huanchen Zhang, Alvin Cheung, Joseph Gonzalez, et~al.
\newblock Continuum: Efficient and robust multi-turn llm agent scheduling with kv cache time-to-live.
\newblock {\em arXiv preprint arXiv:2511.02230}, 2025.

\bibitem{liu2024cachegen}
Yuhan Liu, Hanchen Li, Yihua Cheng, Siddhant Ray, Yuyang Huang, Qizheng Zhang, Kuntai Du, Jiayi Yao, Shan Lu, Ganesh Ananthanarayanan, et~al.
\newblock Cachegen: Kv cache compression and streaming for fast large language model serving.
\newblock In {\em Proceedings of the ACM SIGCOMM 2024 Conference}, pages 38--56, 2024.

\bibitem{10.1145/3689031.3696098}
Jiayi Yao, Hanchen Li, Yuhan Liu, Siddhant Ray, Yihua Cheng, Qizheng Zhang, Kuntai Du, Shan Lu, and Junchen Jiang.
\newblock Cacheblend: Fast large language model serving for rag with cached knowledge fusion.
\newblock In {\em Proceedings of the Twentieth European Conference on Computer Systems}, page 94–109, 2025.

\bibitem{cheng2025lmcache}
Yihua Cheng, Yuhan Liu, Jiayi Yao, Yuwei An, Xiaokun Chen, Shaoting Feng, Yuyang Huang, Samuel Shen, Kuntai Du, and Junchen Jiang.
\newblock Lmcache: An efficient {KV} cache layer for enterprise-scale {LLM} inference.
\newblock {\em CoRR}, abs/2510.09665, 2025.

\bibitem{servegen}
Yuxing Xiang, Xue Li, Kun Qian, Yan Zhang, Wenyuan Yu, Ennan Zhai, Xin Jin, and Jingren Zhou.
\newblock $\{$ServeGen$\}$: Workload characterization and generation of large language model serving in production.
\newblock In {\em 23rd USENIX Symposium on Networked Systems Design and Implementation (NSDI 26)}, pages 1845--1859, 2026.

\bibitem{castillo-lopez-etal-2025-intent}
Galo Castillo-L{\'o}pez, Gael de~Chalendar, and Nasredine Semmar.
\newblock Intent recognition and out-of-scope detection using {LLM}s in multi-party conversations.
\newblock In Fr{\'e}d{\'e}ric B{\'e}chet, Fabrice Lef{\`e}vre, Nicholas Asher, Seokhwan Kim, and Teva Merlin, editors, {\em Proceedings of the 26th Annual Meeting of the Special Interest Group on Discourse and Dialogue}, pages 504--512, Avignon, France, August 2025. Association for Computational Linguistics.

\bibitem{10.1145/3637528.3671647}
Mengting Wan, Tara Safavi, Sujay~Kumar Jauhar, Yujin Kim, Scott Counts, Jennifer Neville, Siddharth Suri, Chirag Shah, Ryen~W. White, Longqi Yang, Reid Andersen, Georg Buscher, Dhruv Joshi, and Nagu Rangan.
\newblock Tnt-llm: Text mining at scale with large language models.
\newblock In {\em Proceedings of the 30th ACM SIGKDD Conference on Knowledge Discovery and Data Mining}, KDD '24, page 5836–5847, New York, NY, USA, 2024. Association for Computing Machinery.

\bibitem{sharegpt}
anon8231489123.
\newblock Sharegpt dataset in huggingface, 2023.

\bibitem{zheng2023lmsys}
Lianmin Zheng, Wei-Lin Chiang, Ying Sheng, Tianle Li, Siyuan Zhuang, Zhanghao Wu, Yonghao Zhuang, Zhuohan Li, Zi~Lin, Eric~P Xing, et~al.
\newblock Lmsys-chat-1m: A large-scale real-world llm conversation dataset.
\newblock {\em arXiv preprint arXiv:2309.11998}, 2023.

\bibitem{agarwal2025gpt}
Sandhini Agarwal, Lama Ahmad, Jason Ai, Sam Altman, Andy Applebaum, Edwin Arbus, Rahul~K Arora, Yu~Bai, Bowen Baker, Haiming Bao, et~al.
\newblock gpt-oss-120b \& gpt-oss-20b model card.
\newblock {\em arXiv preprint arXiv:2508.10925}, 2025.

\bibitem{adamw}
Ilya Loshchilov and Frank Hutter.
\newblock Decoupled weight decay regularization.
\newblock In {\em International Conference on Learning Representations}, 2019.

\bibitem{sglang_github}
{SGLang Project}.
\newblock {SGLang: A fast serving framework for large language models and vision language models}.
\newblock \url{https://github.com/sgl-project/sglang/tree/v0.5.3}, 2024.
\newblock GitHub repository, version v0.5.3, Apache-2.0 license, accessed April 20, 2026.

\bibitem{grattafiori2024llama}
Aaron Grattafiori, Abhimanyu Dubey, Abhinav Jauhri, Abhinav Pandey, Abhishek Kadian, Ahmad Al-Dahle, Aiesha Letman, Akhil Mathur, Alan Schelten, Alex Vaughan, et~al.
\newblock The llama 3 herd of models.
\newblock {\em arXiv preprint arXiv:2407.21783}, 2024.

\bibitem{yang2025qwen3}
An~Yang, Anfeng Li, Baosong Yang, Beichen Zhang, Binyuan Hui, Bo~Zheng, Bowen Yu, Chang Gao, Chengen Huang, Chenxu Lv, et~al.
\newblock Qwen3 technical report.
\newblock {\em arXiv preprint arXiv:2505.09388}, 2025.

\bibitem{qwen2024qwen15}
Jinze Bai, Shuai Bai, Yunfei Chu, Zeyu Cui, Kai Dang, Xiaodong Deng, Yang Fan, Wenbin Ge, Yu~Han, Fei Huang, et~al.
\newblock Qwen technical report.
\newblock {\em arXiv preprint arXiv:2309.16609}, 2023.

\bibitem{wang2024multilingual}
Liang Wang, Nan Yang, Xiaolong Huang, Linjun Yang, Rangan Majumder, and Furu Wei.
\newblock Multilingual e5 text embeddings: A technical report.
\newblock {\em arXiv preprint arXiv:2402.05672}, 2024.

\bibitem{modernbert}
Benjamin Warner, Antoine Chaffin, Benjamin Clavi{\'e}, Orion Weller, Oskar Hallstr{\"o}m, Said Taghadouini, Alexis Gallagher, Raja Biswas, Faisal Ladhak, Tom Aarsen, Griffin~Thomas Adams, Jeremy Howard, and Iacopo Poli.
\newblock Smarter, better, faster, longer: A modern bidirectional encoder for fast, memory efficient, and long context finetuning and inference.
\newblock In Wanxiang Che, Joyce Nabende, Ekaterina Shutova, and Mohammad~Taher Pilehvar, editors, {\em Proceedings of the 63rd Annual Meeting of the Association for Computational Linguistics (Volume 1: Long Papers)}, pages 2526--2547, Vienna, Austria, July 2025. Association for Computational Linguistics.

\bibitem{lpc_github}
{Yangdsh}.
\newblock {LPC: Learned LLM Prefix Cache}.
\newblock \url{https://github.com/yangdsh/LPC}, 2025.
\newblock GitHub repository. Accessed: 2026-05-05.

\end{thebibliography}

 %%%%%%%%%%%%%%%%%%%%%%%%%%%%%%%%%%%%%%%%%%%%%%%%%%%%%%%%%%%%%%%%%%%%%%%%%%%%%%%%%%%%%%%%%%%%%
\newpage
\appendix

\section{Implementation and Experiment Details}

\subsection{Training Intent Classifier}
\label{app:intent_classifier_training}

Elaborating on Section~\ref{sec:intent_classifier}, we train the intent classifier using the labeled conversations from the offline training dataset. For each conversation, we use up to the first four turns as input. At serving, the classifier uses up to the last four turns of the current conversation (e.g., turn 5 uses turns 2--5), following LPC~\cite{yanglearned}. Each turn is marked with a turn index, and the full input is truncated to a maximum length of 1024 tokens. We add the query prefix used by the embedding model and apply head-tail truncation within each turn to preserve both the beginning and ending context of long requests.

We use a frozen embedding backbone to avoid expensive end-to-end fine-tuning. Given the tokenized input, the embedding model produces hidden states, which are mean-pooled over non-padding tokens and then $\ell_2$-normalized. These normalized embeddings are precomputed and cached before training the classifier head. This makes classifier training lightweight and isolates the cost of the embedding model from the MLP optimization.

The classifier head is a small MLP with two hidden layers:
$\text{LayerNorm} \rightarrow \text{Linear}(d_{\mathrm{emb}},2048) \rightarrow \text{GELU} \rightarrow \text{Dropout} \rightarrow \text{Linear}(2048,1024) \rightarrow \text{GELU} \rightarrow \text{Dropout} \rightarrow \text{Linear}(1024,K)$, where $K$ is the number of intent classes and $d_{\mathrm{emb}}$ is the embedding dimension, e.g., 1024 for E5-Large-Inst. In the final system, we use $K=30$.

We split the labeled data into training, validation, and test sets using an 80/10/10 stratified split. The MLP is trained with cross-entropy loss using AdamW with learning rate $3\times10^{-3}$, weight decay $0.01$, dropout $0.25$, and a cosine learning-rate schedule with 5\% warmup. We train for up to 25 epochs and select the checkpoint with the best validation macro-F1 score. Final classifier accuracy is reported on the held-out test set.

\subsection{Implementation of IAC}
\label{app:implementation}

We implement IAC on top of SGLang (v0.5.3), an open-source LLM serving system with a modular runtime. SGLang manages its KV cache using a RadixTree, where each node stores a cached token-prefix segment and shared prefixes are stored only once. This structure naturally supports prefix reuse and makes SGLang a suitable platform for implementing prefix-cache eviction policies.

We extend each RadixTree node with the metadata required by IAC: the predicted intent label, turn index, creation timestamp, and cached-prefix size. When a node is created, IAC assigns its intent label using the request conversation context,
\[
\texttt{node.intent} = \texttt{IntentClassifier}(\texttt{Req.conversation}).
\]
During eviction, SGLang considers leaf nodes as eviction candidates. IAC replaces the default priority with our intent-aware priority score. For each candidate node, we compute its age as
\[
\texttt{node.age} = \texttt{system.clock} - \texttt{node.timestamp},
\]
and then compute the priority from the node's intent, turn index, age, and size using Eq.~\eqref{eq:iac_score}. Nodes with lower priority are evicted first. This integration keeps eviction lightweight and fits directly into SGLang's existing heap-based eviction pipeline.

To support statistics estimation, IAC maintains lightweight request logs. Each record stores the session ID, request ID, turn index, timestamp, and intent label. If a request does not receive a follow-up within 30 minutes, we treat the corresponding state as completed. IAC then estimates continuation probability and delay statistics for each intent--turn state from the recent-logs. These updates require only simple counts and delay summaries, so the update cost remains small.

To handle sparse workload states, IAC uses default statistics when empirical estimates are unreliable. If a state \((i,t)\) is unobserved in the recent-log window, or if \(C(i,t)=0\), we do not compute \(C(i,t+1)/C(i,t)\). Instead, we set \(P(t+1 \mid i,t)=0.5\) and use a default median reuse window of \(150\,\mathrm{s}\). These defaults avoid division-by-zero errors and assign a neutral priority to rare intents or deep conversational turns. Once sufficient observations are available, IAC replaces the defaults with the empirical continuation probability and learned delay statistics for that state.

\subsection{Implementation of Baseline}
\label{app:baselines}

\paragraph{Learned Prefix Caching (LPC) policy.} We implement Learned Prefix Caching (LPC)~\cite{yanglearned} following the authors' open-source implementation and training recipe~\cite{lpc_github}. LPC trains a predictor that takes the conversation history as input and estimates the probability that the conversation will continue. Following the LPC design, we use the most recent user prompts, i.e., the current prompt and up to four preceding prompts, as input context, keep the embedding backbone frozen, and train only the MLP predictor for each workload. During eviction, we use the LPC priority score with its exponential time-decay function, which lowers the predicted continuation probability as the cached prefix becomes inactive. LPC is trained separately for each workload using the common training split described in Appendix~\ref{app:experiment_setup}. We select the best checkpoint based on validation macro-F1 and evaluate LPC on the same replay trace used for all policies.

\paragraph{Workload-Aware (WA) policy.} We adapt WA to SGLang's RadixTree-based prefix cache. In vLLM-style block caches, the offset term prevents earlier prefix blocks from being evicted before later dependent blocks. In SGLang, eviction candidates are RadixTree leaf nodes, and prefix dependencies are already preserved because internal prefix nodes are not evicted before their descendants. Therefore, we set \(\texttt{Offset}=0\) for all candidates and use the workload-specific \(\texttt{ReuseProb}\) as the WA priority. WA is trained and evaluated using the common split described in Appendix~\ref{app:experiment_setup}.

\subsection{Experiment Details}
\label{app:experiment_setup}

\begin{table}[t]
    \caption{Summary of the datasets used in this study. We use ShareChat and WildChat as the main workloads because they provide original user timestamps. LMSys-Chat and ShareGPT are used to evaluate different request arrival patterns, while all four datasets are used to train the intent classifier.}
  \label{tab:workloads}
  \centering
  \begin{tabular}{lccccc}
    \toprule
    Dataset & Timestamp & \#Samples & Avg. length & Avg. turns & \#Languages \\
    \midrule
    ShareChat~\cite{yan2025sharechat}       & \cmark & 142,808   & 1250 & 4.62 & 101 \\
    WildChat~\cite{zhao2024wildchat}        & \cmark & 1,039,785 & 736  & 2.54 & 68  \\
    LMSys-Chat~\cite{zheng2023lmsys}  & \xmark & 1,000,000 & 284  & 2.02 & 65  \\
    ShareGPT~\cite{sharegpt}        & \xmark & 94,145    & 442 & 3.51 & 41  \\
    \bottomrule
  \end{tabular}
\end{table}

\paragraph{Workload split.}
Table~\ref{tab:workloads} presents the summary of the datasets used in this study. To ensure fair data access across learned policies, we use the same history and replay split for all methods. For WildChat and ShareChat, we sort conversations by timestamp. For LMSys-Chat and ShareGPT, which do not provide original timestamps, we use the original dataset order. For each workload, the first 40{,}000 conversations are used as the history/training split, and the next 10{,}000 conversations are used as the replay trace.

\paragraph{Training and history window.}
IAC is designed to update lightweight statistics from completed request logs during deployment. However, in the main evaluation, we use the 40{,}000-conversation training split as the recent-log window for IAC. This matches the information available to the offline learned baselines and avoids giving IAC additional information from the replay trace. WA and LPC are also trained using the same 40{,}000-conversation split. Thus, all learned methods start from the same historical information and are evaluated on the same future 10{,}000-conversation trace.

\paragraph{Replay setting.}
During replay, all policies use the same request order, cache capacity, model, hardware setup, and arrival process. For WildChat and ShareChat, which provide real timestamps, we preserve the original inter-turn delays within each conversation. We also preserve the original conversation order in the workload, but control the relative arrival time between new conversations using a Poisson process at the target arrival rate. This allows us to match the offered load to our testbed while retaining realistic timing within each conversation. For example, replaying 10{,}000 new conversations at an arrival rate of 5 new conversations/s takes approximately 2{,}000 seconds, or 33.3 minutes. Since each conversation has multiple turns, e.g., roughly 3 turns on average, this corresponds to about 30{,}000 total requests and an average request rate of approximately \(5 \times 3 = 15\) requests/s. Follow-up turns are scheduled according to the original inter-turn delays. For synthetic workloads from LMSys-Chat and ShareGPT, which do not provide timestamps, we sample inter-turn delays from a long-tailed delay distribution following prior workload studies~\cite{servegen,wang2025kvcache}.

\section{Additional Experiments}

\begin{figure}[t]
    \centering
    \resizebox{0.7\textwidth}{!}{%
    \begin{minipage}{\textwidth}

        \begin{subfigure}[t]{0.48\linewidth}
            \centering
            \includegraphics[width=\linewidth]{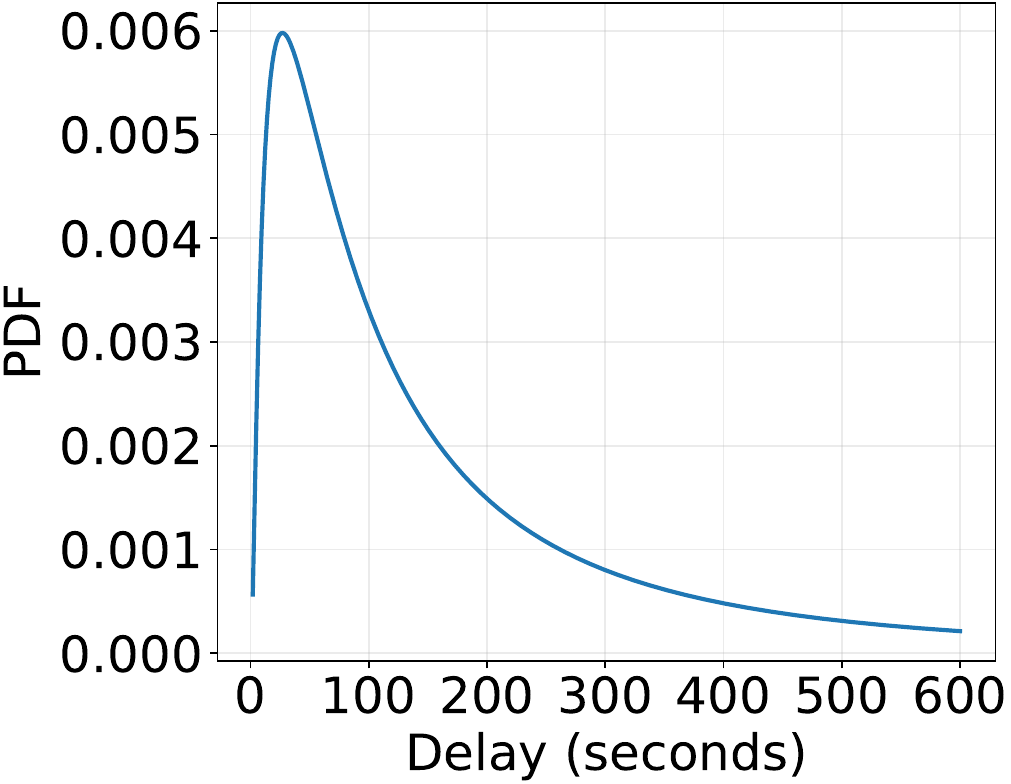}
            \caption{PDF of delay}
            \label{fig:pdf_of_delay}
        \end{subfigure}
        \hfill
        \begin{subfigure}[t]{0.48\linewidth}
            \centering
            \includegraphics[width=\linewidth]{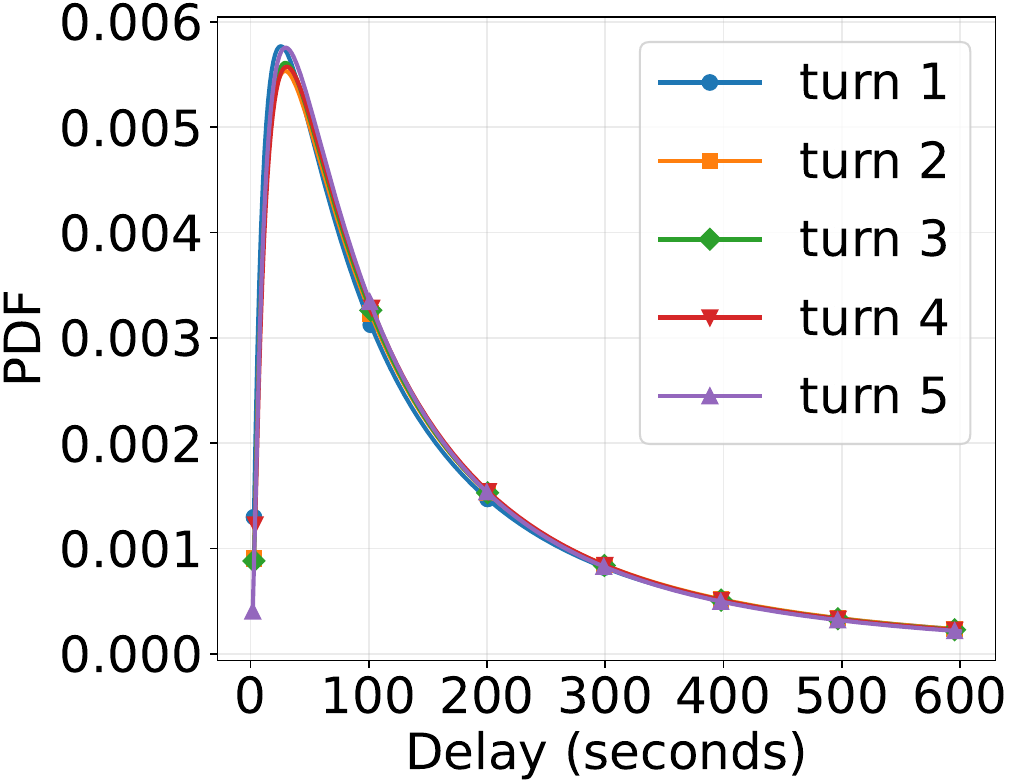}
            \caption{PDF of delay by turn}
            \label{fig:pdf_of_delay_by_turn}
        \end{subfigure}

        \vspace{0.5em}

        \begin{subfigure}[t]{0.48\linewidth}
            \centering
            \includegraphics[width=\linewidth]{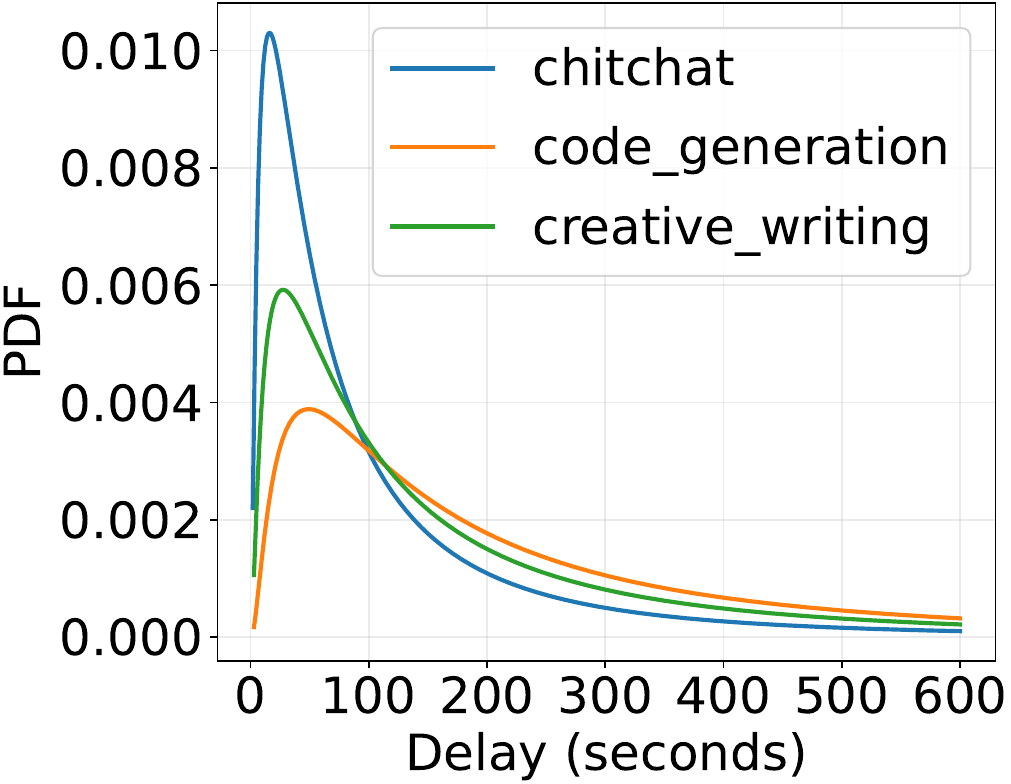}
            \caption{PDF of delay by intent}
            \label{fig:pdf_of_delay_by_intnet}
        \end{subfigure}
        \hfill
        \begin{subfigure}[t]{0.48\linewidth}
            \centering
            \includegraphics[width=\linewidth]{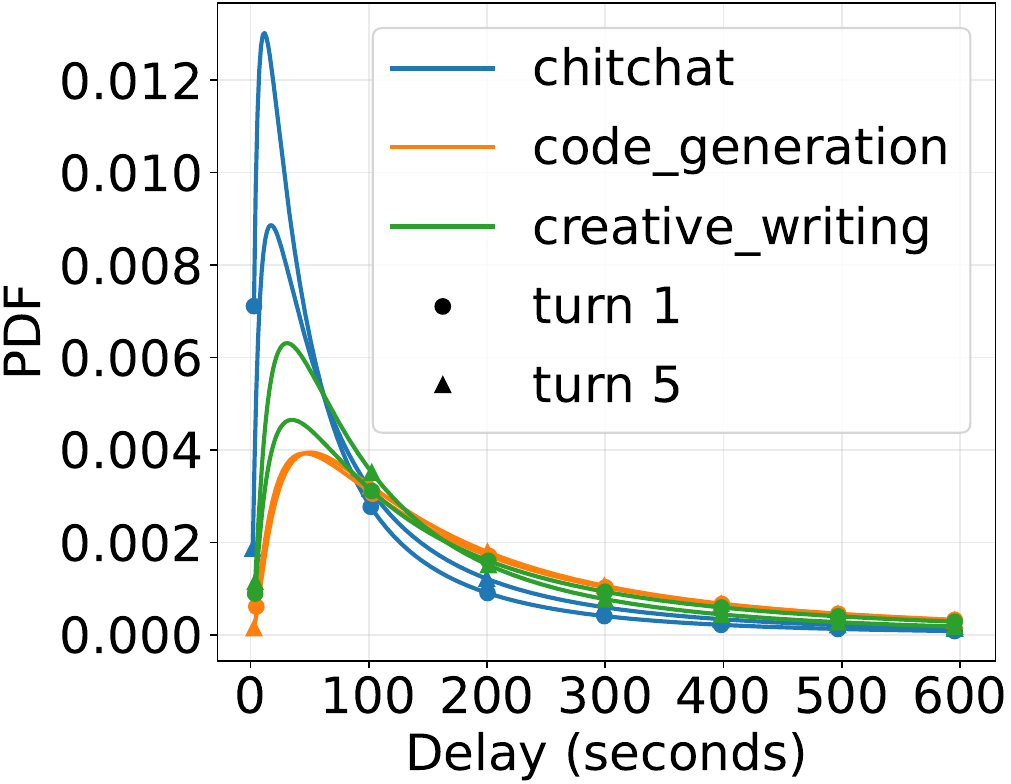}
            \caption{PDF of delay by intent and turn}
            \label{fig:pdf_by_intent_turn}
        \end{subfigure}

    \end{minipage}%
    }

    \caption{
    Inter-turn delay distributions on WildChat-1M~\cite{zhao2024wildchat}. 
    Progressively finer grouping by workload signals exposes cache-reuse structure that is hidden by a single global distribution, motivating our design. 
    Although turn index alone provides limited separation for delay modeling, it remains useful for estimating reuse probability.
    }
    \label{fig:delay_pdf_all}
\end{figure}

\subsection{Inter-turn delay distributions}
\label{app:additional_insights}
Fig.~\ref{fig:delay_pdf_all} further illustrates how delay patterns become more structured under finer workload groupings. When all requests are aggregated together, the delay distribution appears broad and smooth. Grouping by turn alone reveals only limited variation. In contrast, grouping by intent exposes clear differences in delay behavior across workload types, and combining intent with turn reveals even finer-grained patterns. This further supports our design choice of modeling cache reuse at the \((\textit{intent}, \textit{turn})\) level.

\subsection{Overhead of assigning the intent}
\label{app:overhead}

The intent classifier assigns the intent label as soon as a request arrives in the scheduler, while the request waits in the queue. Therefore, the classification overlaps with queuing, and its worst-case overhead on the critical path is the 20\,ms classifier latency when there is no queuing delay.

Table~\ref{tab:batching} shows that batching can amortize the cost. At 100 requests/s, a single instance needs only about 0.2\,ms per request at a batch size of 100 and the peak GPU memory of the classifier stays 2.29\,GB across batch sizes.

\begin{table}[h]
\centering
\caption{ Latency, throughput, and memory of the E5-Large-Inst intent classifier with batching.}
\label{tab:batching}
\small
 \begin{tabular}{rccccc}
\toprule
Batch size & p50 (ms) & p95 (ms) & Per-request (ms) & Throughput (req/s) & Peak GPU mem. (GB) \\
\midrule
1     & 19.4 & 20.6 & 19.4  & 52     & 2.29 \\
10    & 19.4 & 19.8 & 1.94  & 515    & 2.29 \\
100   & 19.6 & 19.9 & 0.20  & 5,112  & 2.29 \\
\bottomrule
\end{tabular}
\end{table}

\subsection{Online Serving with IAC}
\label{app:online_serving}

In the main evaluation, IAC estimates its statistics once from the 40K-conversation history split and is then evaluated on the following 10K conversations (Appendix~\ref{app:experiment_setup}). In deployment, however, IAC is meant to run online: it periodically re-estimates its statistics from recent request logs as the workload drifts (Section~\ref{sec:online_learning}). To check that IAC works in this setting, we run a causal rolling-window experiment that mimics online serving.

We split the WildChat stream into sequential 10K-conversation folds and use a cache capacity of 140K tokens. The first fold (0--10K) provides the initial statistics, and the next four folds (10K--50K) are replayed in order. Before each fold, IAC re-estimates the \((i,t)\) statistics using only past folds, with a window of \(N=10\)K conversations, so no future information is used. As shown in Table~\ref{tab:rolling_full}, IAC keeps a clear advantage over LRU in every fold (+49\% to +83\%), even though the statistics are refreshed from a much smaller window than in the main evaluation.

Updating the statistics requires no separate effort. The inputs (i.e., intent label, turn index, and arrival time of each request) are already recorded in the request logs as requests are served, so no extra data collection or labeling is needed. Each update is a single pass that counts continuations and fits the log-normal delay parameters for every \((i,t)\) state, with no model training. It takes about 1.4\,s on CPU and runs off the serving path, so IAC can refresh its statistics frequently at negligible cost, unlike ML-based policies that must be retrained to track drift.

\begin{table}[h]
\centering
\caption{ Causal rolling-window evaluation on WildChat (140K token capacity, \(N=10\)K). The statistics are re-estimated from past folds only. Cache hit rate in \%.}
\label{tab:rolling_full}
\small
 \begin{tabular}{lcccc}
\toprule
Replay window & LRU & IAC (ours) & Gain & Update time (CPU) \\
\midrule
10K--20K & 13.7 & 20.4 & +49\% & 1.32\,s \\
20K--30K & 9.8  & 17.9 & +83\% & 1.41\,s \\
30K--40K & 14.2 & 23.5 & +66\% & 1.37\,s \\
40K--50K & 11.6 & 19.7 & +70\% & 1.53\,s \\
\bottomrule
\end{tabular}
\end{table}

\subsection{Using the Full Delay Distribution}
We test a variant, called Full-Horizon, that replaces the binary recency test \(\mathbf{1}[\mathrm{age}<M_{i,t}]\) with a continuous score, i.e., the probability that the next turn arrives within the next \(H\) seconds given the current age of the entry, computed from the learned log-normal delay model. Full-Horizon uses this probability as the primary signal. As shown in Table~\ref{tab:fullhorizon}, Full-Horizon stays below IAC at every capacity on ShareChat for two reasons. First, IAC keeps a binary recency threshold like the TB+S oracle it mimics, so recency, reuse probability, and size act in a clean order. A continuous recency score is almost unique per entry, so it dominates the ranking, and the reuse-probability and size signals rarely take effect. Second, Full-Horizon relies on the full delay distribution, including its spread and tail, which is noisy per \((i,t)\) state, whereas the median is robust. This also matches Fig.~\ref{fig:mean_meadian_mode_ablation}, where the median outperforms the mean and the mode.

\begin{table}[h]
\centering
\caption{Cache hit rate (\%) on ShareChat with the binary median-based recency test (IAC) and with the full delay distribution (Full-Horizon).}
\label{tab:fullhorizon}
\small
\begin{tabular}{ccc}
\toprule
Capacity (\(\times 10^3\) tokens) & Full-Horizon & IAC (ours) \\
\midrule
80  & 21.5 & 23.7 \\
110 & 27.4 & 29.8 \\
140 & 31.0 & 34.6 \\
170 & 35.3 & 38.5 \\
\bottomrule
\end{tabular}
\end{table}

\subsection{Comparison with WA}
\label{app:wa_results}

\begin{wrapfigure}{r}{0.27\textwidth}
    \centering
    \vspace{-3em}
    \includegraphics[width=\linewidth]{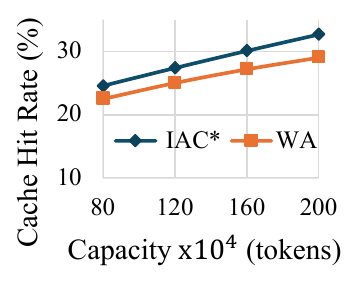}
    \caption{WA~\cite{wang2025kvcache} trace.}
    \vspace{-1em}
    \label{fig:wa_cache_hit}
\end{wrapfigure}

We further compare with workload-aware (WA)~\cite{wang2025kvcache} separately, as its production trace exposes only hashed inputs with four coarse request types: text, multimodal, file, and search with timestamps. Since hashing prevents intent inference, we replace IAC's 30 intent categories with these four request types and call this variant IAC*. As shown in Fig.~\ref{fig:wa_cache_hit}, IAC* outperforms WA even though both policies group requests identically, showing that IAC's eviction design remains effective even with coarse request-type signals.

In addition, we test the WA policy on our workloads. WA estimates reuse probability and spatial locality from training data, grouped by request type (text, multimodal, file, and search) and turn index. Since our primary workloads contain only text requests, WA can only use turn-level categories in this setting. In its original block-based cache design for vLLM~\cite{vllm}, WA prioritizes each KV block using a tuple based on \(\texttt{ReuseProb}\) and \(\texttt{Offset}\). Here, \(\texttt{ReuseProb}\) is estimated from the fitted reuse distribution of the request category, while \(\texttt{Offset}\) gives higher priority to earlier KV blocks to preserve prefix dependencies. 

\begin{figure*}[t]
    \centering

    \begin{subfigure}[t]{0.35\textwidth}
        \centering
        \includegraphics[width=\linewidth]{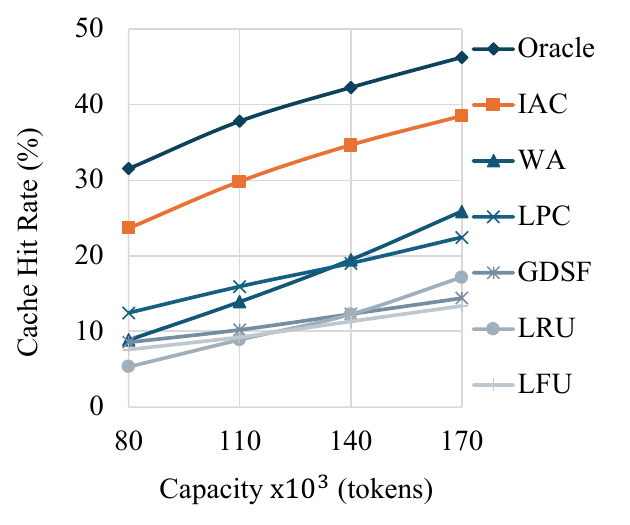}
        \caption{Cache hit rate}
        \label{fig:cache_hit_wa}
    \end{subfigure}
    \hfill
    \begin{subfigure}[t]{0.63\textwidth}
        \centering
        \includegraphics[width=\linewidth]{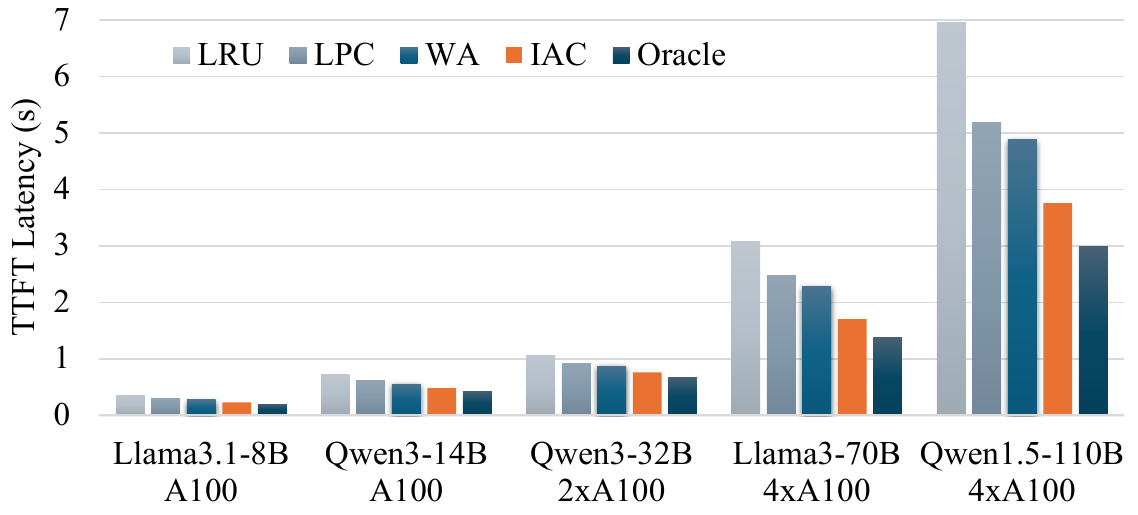}
        \caption{TTFT latency}
        \label{fig:throughput_wa}
    \end{subfigure}

    \caption{
    Extended evaluation including WA on ShareChat.
    (\textbf{a}) Cache hit rate across cache capacities.
    (\textbf{b}) TTFT latency across serving settings.
    }
    \label{fig:combined_eval_wa}
\end{figure*}

Fig.~\ref{fig:cache_hit_wa} and Fig.~\ref{fig:throughput_wa} report cache hit rate and TTFT when WA is included in the comparison. The trend is consistent with the main evaluation in Section~\ref{sec:evaluation}: IAC achieves higher cache hit rate and lower TTFT than WA. For example, on ShareChat at 140K token capacity, IAC reaches 34.6\% cache hit rate, compared with 19.4\% for WA. This improvement also translates to latency reduction. On Qwen-1.5-110B, IAC reduces TTFT to 3.76\,s, compared with 5.90\,s for WA. Similarly, on Llama-3-70B, IAC reduces TTFT to 1.7\,s, compared with 2.3\,s for WA. These results show that even when WA is approximated on our text-only workloads, IAC remains more effective because it uses finer-grained workload categories and delay-aware reuse characterization.

\subsection{Robustness to Arrival Rates on LMSys-Chat}
\label{app:lmsys}

\begin{wrapfigure}{r}{0.35\textwidth}
    \centering
    \vspace{-1.5em}
    \includegraphics[width=\linewidth]{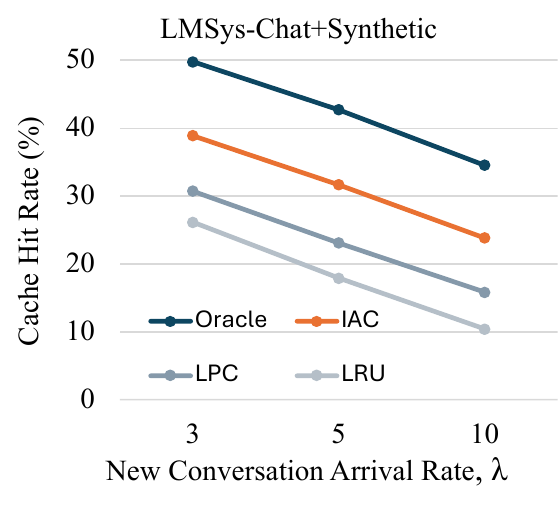}
    \caption{Cache hit rate on LMSys-Chat under different arrival rates \(\lambda\) (150K tokens).}
    \label{fig:lmsys_hit_rate}
    \vspace{-1.5em}
\end{wrapfigure}

To complement the ShareGPT results in Section~\ref{sec:cachehitrate} (Fig.~\ref{fig:sharegpt_hit_rate}), we evaluate IAC on LMSys-Chat~\cite{zheng2023lmsys}, which also lacks original timestamps. We construct a synthetic timestamped workload in the same way: new conversations arrive following a Poisson process with rate \(\lambda\), while inter-turn delays within each conversation are sampled from a long-tailed distribution with mean 4.6 and variance 1, following prior workload studies~\cite{servegen}. The cache capacity is fixed at 150K tokens.

Fig.~\ref{fig:lmsys_hit_rate} shows the cache hit rate as \(\lambda\) increases. Higher arrival rates make the workload less cache-friendly for all policies, because more new conversations compete for cache space and force earlier eviction of prefixes from ongoing conversations. IAC remains consistently better than the baselines across all arrival rates. However, its gains are smaller than on the real timestamped workloads (WildChat and ShareChat), because all intent categories share the same synthetic delay distribution. This removes most of the differences in reuse delay across intents and thus limits the benefit of IAC's delay-aware recency signal, leaving the reuse-probability and size signals to drive the improvement.

\subsection{Robustness to Arrival Patterns and Uniform Delays}
We test bursty and diurnal arrivals on WildChat by sampling the arrival times of new conversations from ON/OFF, Pareto, and diurnal patterns. Table~\ref{tab:arrival} shows that IAC keeps its advantage in all cases, because its delay-aware signal is learned from inter-turn delays, which the arrival process of new conversations does not change. We further replace all WildChat inter-turn delays with a constant 100\,s (Table~\ref{tab:uniform}). IAC still far exceeds LRU and stays close to the oracle. With a fixed 100\,s delay, every state has \(M_{i,t}=100\)\,s, so the reuse window is known exactly and the recency test has no estimation error. IAC then orders entries by continuation probability and size, which is enough to track the oracle. A uniform delay therefore does not remove the benefit of IAC. It only shifts the gain from the delay signal to the reuse-probability signal.

\begin{table}[h]
\centering
\begin{minipage}[t]{0.45\linewidth}
\centering
\caption{Cache hit rate (\%) on WildChat at 140K token capacity under different arrival patterns of new conversations.}
\label{tab:arrival}
\small
\begin{tabular}{lccc}
\toprule
Arrival pattern & LRU & IAC (ours) & Oracle \\
\midrule
Diurnal & 12.1 & 19.7 & 28.5 \\
Pareto  & 12.5 & 22.7 & 28.4 \\
ON/OFF  & 14.4 & 22.5 & 32.2 \\
\bottomrule
\end{tabular}
\end{minipage}
\hfill
\begin{minipage}[t]{0.5\linewidth}
\centering
\caption{Cache hit rate (\%) on WildChat when every inter-turn delay is set to a constant 100\,s.}
\label{tab:uniform}
\small
\begin{tabular}{cccc}
\toprule
Capacity & LRU & IAC (ours) & Oracle \\
\midrule
80 \(\times 10^3\)  & 2.3 & 9.3  & 13.0 \\
110 \(\times 10^3\) & 2.7 & 12.8 & 17.2 \\
140 \(\times 10^3\) & 3.2 & 16.7 & 21.4 \\
170 \(\times 10^3\) & 3.9 & 20.2 & 25.1 \\
\bottomrule
\end{tabular}
\end{minipage}
\end{table}

\subsection{Tail Latency}
\label{app:tail_latency}
Table~\ref{tab:tail} extends Fig.~\ref{fig:ttft_latency} with the p50 (median) and p90 TTFT on Llama-3-70B (4\(\times\)A100) at 10 requests/s. Since IAC raises the cache hit rate and reduces recomputation, it also improves the tail.

\begin{table}[h]
\centering
\caption{Mean and tail TTFT (s) on Llama-3-70B (4\(\times\)A100) at 10 requests/s.}
\label{tab:tail}
\small
\begin{tabular}{lccc}
\toprule
Policy & Mean & p50 & p90 \\
\midrule
LRU        & 6.51 & 4.29 & 17.80 \\
LPC        & 4.91 & 3.67 & 11.84 \\
IAC (ours) & 3.43 & 3.03 & 6.77 \\
Oracle     & 3.21 & 2.07 & 5.06 \\
\bottomrule
\end{tabular}
\end{table}

\subsection{Per-Request Cache Hit Ratio}
Table~\ref{tab:cdf} reports the distribution of the per-request cache hit ratio at 16 requests/s and 170K token capacity. The ratio is 0 at the low percentiles under every policy, because single-turn requests and first turns have no cached prefix to reuse. The gains of IAC appear in the upper percentiles, where reuse is possible.

\begin{table}[h]
\centering
\caption{Percentiles of the per-request cache hit ratio (\%) at 16 requests/s and 170K token capacity. Overall is the token-weighted cache hit rate used in the rest of the paper.}
\label{tab:cdf}
\small
\begin{tabular}{lcccccc}
\toprule
Policy & p10 & p25 & p50 & p75 & p90 & Overall \\
\midrule
LRU        & 0.0 & 0.0 & 0.6 & 13.3 & 75.4 & 17.1 \\
LPC        & 0.0 & 0.0 & 0.7 & 12.8 & 74.2 & 22.4 \\
IAC (ours) & 0.0 & 0.0 & 0.3 & 14.3 & 86.8 & 38.5 \\
Oracle     & 0.0 & 0.0 & 0.1 & 7.7  & 69.9 & 46.2 \\
\bottomrule
\end{tabular}
\end{table}

\section{Intent Taxonomy and Labeling Prompt}
\label{app:intend_discover_pipeline}

We present the intent taxonomy used in our workload analysis. We first list the 30-category label space and then present the 20-category and 10-category variants by grouping the original labels under each merged category.

\subsection{Thirty Intent Categories}
\label{app:intent_30}

\begin{enumerate}
    \item \texttt{general\_explanations\_definitions}: Broad factual explanations, concept definitions, or general knowledge questions.
    \item \texttt{creative\_writing}: Creative writing such as stories, poems, scripts, or imaginative text generation.
    \item \texttt{math\_problem\_solving}: Solving math questions, equations, symbolic problems, or numerical calculations.
    \item \texttt{writing\_business\_marketing}: Business communication, professional writing, ads, pitches, or marketing content.
    \item \texttt{code\_generation}: Writing code from instructions, generating scripts, or implementing programming solutions.
    \item \texttt{how\_to\_guides}: Step-by-step instructions for completing a task or learning how to do something.
    \item \texttt{relationship}: Questions or discussion about romantic or interpersonal relationship issues.
    \item \texttt{academic\_writing}: Academic writing such as essays, reports, research text, or scholarly-style content.
    \item \texttt{health\_medical\_info}: Health, symptom, medical, fitness, or wellness-related information requests.
    \item \texttt{summarization}: Summarizing text, documents, articles, notes, or conversations.
    \item \texttt{recommendations\_products}: Requests for product suggestions, comparisons, or purchase recommendations.
    \item \texttt{code\_debugging}: Finding, explaining, or fixing bugs, errors, or technical issues in code or systems.
    \item \texttt{chitchat}: Casual conversation, greetings, or light social interaction without a clear task goal.
    \item \texttt{writing\_editing\_proofreading}: Revising, proofreading, shortening, polishing, or improving existing text.
    \item \texttt{data\_analysis\_statistics}: Statistical analysis, data interpretation, quantitative reasoning, or results explanation.
    \item \texttt{entertainment\_media}: Movies, TV, music, books, games, celebrities, or general entertainment topics.
    \item \texttt{career\_jobs}: Career advice, resume help, interview preparation, job search, or workplace questions.
    \item \texttt{legal\_info}: Questions about laws, legal procedures, rights, or legal concepts.
    \item \texttt{data\_extraction\_structuring}: Extracting structured information from unstructured text or reorganizing content into tables, fields, or formats.
    \item \texttt{translation}: Translating content between languages.
    \item \texttt{personal\_life\_advice}: Advice about everyday personal decisions, habits, routines, or general life situations.
    \item \texttt{mental\_wellbeing\_support}: Emotional support, stress-related discussion, motivation, or general mental well-being guidance.
    \item \texttt{research\_deep\_dive}: In-depth analytical or research-oriented requests requiring detailed explanation or synthesis.
    \item \texttt{philosophy\_religion}: Questions about beliefs, ethics, religion, spirituality, or philosophical ideas.
    \item \texttt{travel\_planning}: Planning trips, itineraries, transportation, destinations, or travel logistics.
    \item \texttt{education\_study\_help}: General studying support, concept learning, homework help, or academic tutoring.
    \item \texttt{personal\_finance}: Budgeting, saving, investing, loans, taxes, or other personal finance topics.
    \item \texttt{food\_cooking\_recipes}: Recipes, cooking instructions, ingredient substitutions, or food preparation advice.
    \item \texttt{current\_events\_politics}: News, politics, public affairs, elections, or recent real-world events.
    \item \texttt{system\_admin\_devops}: System setup, deployment, infrastructure, DevOps, cloud, networking, or server management.
\end{enumerate}

\subsection{Twenty-Category Taxonomy}
\label{app:intent_20}

The 20-category taxonomy is formed by merging semantically related labels from the 30-category taxonomy. For each merged category below, we list the original labels assigned to it.

\begin{enumerate}
    \item \texttt{advice\_support}
    \begin{enumerate}
        \item \texttt{relationship}
        \item \texttt{personal\_life\_advice}
        \item \texttt{mental\_wellbeing\_support}
    \end{enumerate}

    \item \texttt{career\_jobs}
    \begin{enumerate}
        \item \texttt{career\_jobs}
    \end{enumerate}

    \item \texttt{chitchat}
    \begin{enumerate}
        \item \texttt{chitchat}
    \end{enumerate}

    \item \texttt{coding\_and\_systems}
    \begin{enumerate}
        \item \texttt{code\_generation}
        \item \texttt{code\_debugging}
        \item \texttt{system\_admin\_devops}
    \end{enumerate}

    \item \texttt{current\_events\_politics}
    \begin{enumerate}
        \item \texttt{current\_events\_politics}
    \end{enumerate}

    \item \texttt{education\_learning}
    \begin{enumerate}
        \item \texttt{math\_problem\_solving}
        \item \texttt{education\_study\_help}
    \end{enumerate}

    \item \texttt{food\_cooking\_recipes}
    \begin{enumerate}
        \item \texttt{food\_cooking\_recipes}
    \end{enumerate}

    \item \texttt{general\_explanations\_definitions}
    \begin{enumerate}
        \item \texttt{general\_explanations\_definitions}
    \end{enumerate}

    \item \texttt{health\_medical\_info}
    \begin{enumerate}
        \item \texttt{health\_medical\_info}
    \end{enumerate}

    \item \texttt{how\_to\_guides}
    \begin{enumerate}
        \item \texttt{how\_to\_guides}
    \end{enumerate}

    \item \texttt{legal\_info}
    \begin{enumerate}
        \item \texttt{legal\_info}
    \end{enumerate}

    \item \texttt{personal\_finance}
    \begin{enumerate}
        \item \texttt{personal\_finance}
    \end{enumerate}

    \item \texttt{philosophy\_religion}
    \begin{enumerate}
        \item \texttt{philosophy\_religion}
    \end{enumerate}

    \item \texttt{recommendations}
    \begin{enumerate}
        \item \texttt{recommendations\_products}
        \item \texttt{entertainment\_media}
    \end{enumerate}

    \item \texttt{research\_deep\_dive}
    \begin{enumerate}
        \item \texttt{data\_analysis\_statistics}
        \item \texttt{data\_extraction\_structuring}
        \item \texttt{research\_deep\_dive}
    \end{enumerate}

    \item \texttt{summarization}
    \begin{enumerate}
        \item \texttt{summarization}
    \end{enumerate}

    \item \texttt{translation}
    \begin{enumerate}
        \item \texttt{translation}
    \end{enumerate}

    \item \texttt{travel\_planning}
    \begin{enumerate}
        \item \texttt{travel\_planning}
    \end{enumerate}

    \item \texttt{writing\_creation}
    \begin{enumerate}
        \item \texttt{creative\_writing}
        \item \texttt{writing\_business\_marketing}
        \item \texttt{academic\_writing}
    \end{enumerate}

    \item \texttt{writing\_editing\_proofreading}
    \begin{enumerate}
        \item \texttt{writing\_editing\_proofreading}
    \end{enumerate}
\end{enumerate}

\subsection{Ten-Category Taxonomy}
\label{app:intent_10}

The 10-category taxonomy further merges the original labels into broader workload families.

\begin{enumerate}
    \item \texttt{advice\_support}
    \begin{enumerate}
        \item \texttt{relationship}
        \item \texttt{personal\_life\_advice}
        \item \texttt{mental\_wellbeing\_support}
    \end{enumerate}

    \item \texttt{career\_education}
    \begin{enumerate}
        \item \texttt{math\_problem\_solving}
        \item \texttt{career\_jobs}
        \item \texttt{education\_study\_help}
    \end{enumerate}

    \item \texttt{chitchat}
    \begin{enumerate}
        \item \texttt{chitchat}
    \end{enumerate}

    \item \texttt{coding\_debugging}
    \begin{enumerate}
        \item \texttt{code\_generation}
        \item \texttt{code\_debugging}
        \item \texttt{system\_admin\_devops}
    \end{enumerate}

    \item \texttt{data\_work}
    \begin{enumerate}
        \item \texttt{data\_analysis\_statistics}
        \item \texttt{data\_extraction\_structuring}
    \end{enumerate}

    \item \texttt{high\_stakes\_info}
    \begin{enumerate}
        \item \texttt{health\_medical\_info}
        \item \texttt{legal\_info}
        \item \texttt{personal\_finance}
    \end{enumerate}

    \item \texttt{how\_to\_practical}
    \begin{enumerate}
        \item \texttt{how\_to\_guides}
        \item \texttt{food\_cooking\_recipes}
    \end{enumerate}

    \item \texttt{recommendations\_planning}
    \begin{enumerate}
        \item \texttt{recommendations\_products}
        \item \texttt{entertainment\_media}
        \item \texttt{travel\_planning}
    \end{enumerate}

    \item \texttt{research\_news}
    \begin{enumerate}
        \item \texttt{general\_explanations\_definitions}
        \item \texttt{research\_deep\_dive}
        \item \texttt{philosophy\_religion}
        \item \texttt{current\_events\_politics}
    \end{enumerate}

    \item \texttt{writing\_language}
    \begin{enumerate}
        \item \texttt{creative\_writing}
        \item \texttt{writing\_business\_marketing}
        \item \texttt{academic\_writing}
        \item \texttt{summarization}
        \item \texttt{writing\_editing\_proofreading}
        \item \texttt{translation}
    \end{enumerate}
\end{enumerate}

\subsection{System Prompt for Intent Labeling}
\label{app:system_prompt}

\begin{tcolorbox}[
    enhanced,
    breakable,
    colback=gray!5,
    colframe=black!75,
    boxrule=0.5pt,
    arc=2.5mm,
    left=2mm,
    right=2mm,
    top=1.5mm,
    bottom=1.5mm,
    title={System Prompt Used for Intent Labeling}
]
You are an intent classification system for human--LLM conversations. Your task is to assign the \textbf{single best intent category} for the \textbf{last user message} in the conversation. Choose exactly one category from the list below:

\begin{enumerate}
    \item \texttt{chitchat}: casual conversation, greetings, or small talk.
    \item \texttt{personal\_life\_advice}: advice on personal decisions, routines, or life situations.
    \item \texttt{relationship\_dating}: romantic relationships, dating, or interpersonal relationship issues.
    \item \texttt{mental\_wellbeing\_support}: emotional support, stress, motivation, or general mental well-being.
    \item \texttt{health\_medical\_info}: health, symptoms, medical, wellness, or fitness information.
    \item \texttt{legal\_info}: legal questions, laws, rights, or legal procedures.
    \item \texttt{personal\_finance}: saving, investing, loans, taxes, budgeting, or money management.
    \item \texttt{career\_jobs}: jobs, resumes, interviews, workplace, or career advice.
    \item \texttt{education\_study\_help}: studying, tutoring, homework help, or academic learning.
    \item \texttt{math\_problem\_solving}: math problems, equations, or numerical calculations.
    \item \texttt{how\_to\_guides}: step-by-step instructions or practical guidance.
    \item \texttt{recommendations\_products}: product suggestions, comparisons, or buying advice.
    \item \texttt{travel\_planning}: trip planning, itineraries, destinations, or travel logistics.
    \item \texttt{food\_cooking\_recipes}: recipes, cooking, ingredients, or food preparation.
    \item \texttt{entertainment\_media}: movies, music, books, games, celebrities, or media topics.
    \item \texttt{current\_events\_politics}: news, politics, elections, or recent public events.
    \item \texttt{philosophy\_religion}: ethics, religion, spirituality, or philosophy.
    \item \texttt{general\_explanations\_definitions}: factual explanations, definitions, or broad knowledge questions.
    \item \texttt{research\_deep\_dive}: in-depth research, detailed analysis, or deep explanatory requests.
    \item \texttt{creative\_writing}: stories, poems, scripts, or imaginative writing.
    \item \texttt{academic\_writing}: essays, reports, academic drafts, or scholarly writing.
    \item \texttt{writing\_business\_marketing}: business writing, marketing copy, professional messages, or pitches.
    \item \texttt{writing\_editing\_proofreading}: editing, proofreading, rewriting, or improving provided text.
    \item \texttt{summarization}: summarizing text, notes, documents, or conversations.
    \item \texttt{translation}: translating text between languages.
    \item \texttt{code\_generation}: generating code from instructions.
    \item \texttt{code\_debugging}: debugging code, fixing errors, or troubleshooting technical problems.
    \item \texttt{system\_admin\_devops}: deployment, infrastructure, servers, cloud, networking, or DevOps operations.
    \item \texttt{data\_analysis\_statistics}: data analysis, statistics, interpretation, or quantitative reasoning.
    \item \texttt{data\_extraction\_structuring}: extracting structured data or converting text into organized fields or formats.
\end{enumerate}

\textbf{Instructions:}
\begin{itemize}
    \item Read the full conversation for context.
    \item Assign the label that best matches the main goal of the last user message.
    \item If multiple categories seem relevant, choose the one that is most central to the last user request.
    \item Output only the category name exactly as written above.
    \item Do not output any explanation.
\end{itemize}

\textbf{Examples:}
\begin{itemize}
    \item ``Can you write a Python script that reads a CSV and removes duplicate rows?'' $\rightarrow$ \texttt{code\_generation}
    \item ``My JavaScript code throws undefined is not a function. Can you fix it?'' $\rightarrow$ \texttt{code\_debugging}
    \item ``Plan a 5-day trip to Japan for me.'' $\rightarrow$ \texttt{travel\_planning}
    \item ``Please shorten and polish this paragraph from my paper.'' $\rightarrow$ \texttt{writing\_editing\_proofreading}
    \item ``Explain what overfitting means in machine learning.'' $\rightarrow$ \texttt{general\_explanations\_definitions}
\end{itemize}

Conversation to classify:

\texttt{\{conversation\_text\}}
\end{tcolorbox}

\end{document}